\pdfoutput=1

\documentclass[11pt]{article}

\usepackage[]{emnlp2022}

\usepackage{placeins}
\usepackage{xcolor}
\usepackage{times}
\usepackage{latexsym}
\usepackage{url}
\usepackage{mathdots}
\usepackage{yhmath}
\usepackage{cancel}
\usepackage{color}
\usepackage{siunitx}
\usepackage{array}
\usepackage{multirow}
\usepackage{amssymb}
\usepackage{fdsymbol}
\usepackage{tabularx}
\usepackage{booktabs}
\usepackage{svg}
\usepackage{tablefootnote}
\usepackage{cleveref}
\usepackage{makecell}
\usepackage{enumitem}
\usepackage{amsfonts}
\usepackage{pifont}
\usepackage{subfigure}
\usepackage[ruled,vlined,noend]{algorithm2e}

\usepackage[T1]{fontenc}

\usepackage[utf8]{inputenc}

\newcommand{\smallsection}[1]{\noindent\textbf{#1.}}
\newcommand\blfootnote[1]{%
  \begingroup
  \renewcommand\thefootnote{}\footnote{#1}%
  \addtocounter{footnote}{-1}%
  \endgroup
}

\usepackage{xspace}
\newcommand{\our}{\mbox{LOPS}\xspace}
\usepackage{microtype}

\title{\our: Learning Order Inspired Pseudo-Label Selection \\ for Weakly Supervised Text Classification}

\author{
  Dheeraj Mekala$^{\diamondsuit}$ $\qquad$ Chengyu Dong$^{\diamondsuit}$ $\qquad$  Jingbo Shang$^{\spadesuit, *}$ \\
    $^\diamondsuit$ University of California San Diego\\
    $^\spadesuit$ Hal\i c\i o\u glu Data Science Institute, University of California San Diego\\
   \texttt{\{dmekala, cdong, jshang\}@ucsd.edu}
}

\begin{document}
\maketitle

\begin{abstract}
\blfootnote{$*$ Jingbo Shang is the corresponding author.}
Weakly supervised text classification methods typically train a deep neural classifier based on pseudo-labels.
The quality of pseudo-labels
is crucial to final performance but they are inevitably noisy due to their heuristic nature, so selecting the correct ones has a huge potential for performance boost.
One straightforward solution is to select samples based on the softmax probability scores in the neural classifier corresponding to their pseudo-labels.
However, we show through our experiments that such solutions are ineffective and unstable due to the erroneously high-confidence predictions from poorly calibrated models.
Recent studies on the memorization effects of deep neural models suggest that these models first memorize training samples with clean labels and then those with noisy labels.
Inspired by this observation, we propose a novel pseudo-label selection method \our that takes learning order of samples into consideration.
We hypothesize that the learning order reflects the probability of wrong annotation in terms of ranking, and therefore, propose to select the samples that are learnt earlier.
\our can be viewed as a strong performance-boost plug-in to most existing weakly-supervised text classification methods, as confirmed in extensive experiments on four real-world datasets.
\end{abstract}

\section{Introduction}

\begin{figure}[t]
    \subfigure[]{
        \includegraphics[width=0.45\linewidth]{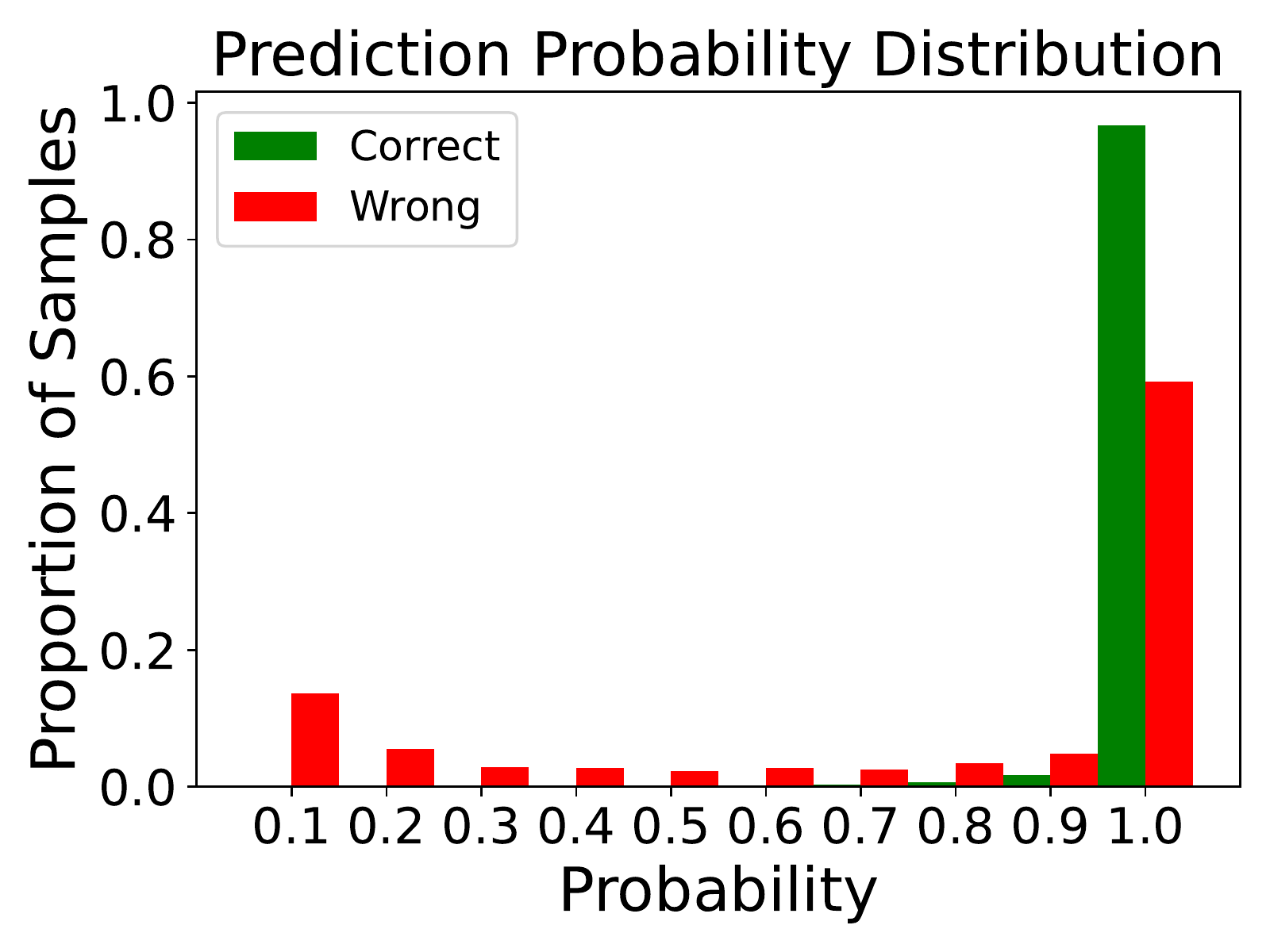}
        \label{figure:prob_filter}
    }
    \subfigure[]{
        \includegraphics[width=0.45\linewidth]{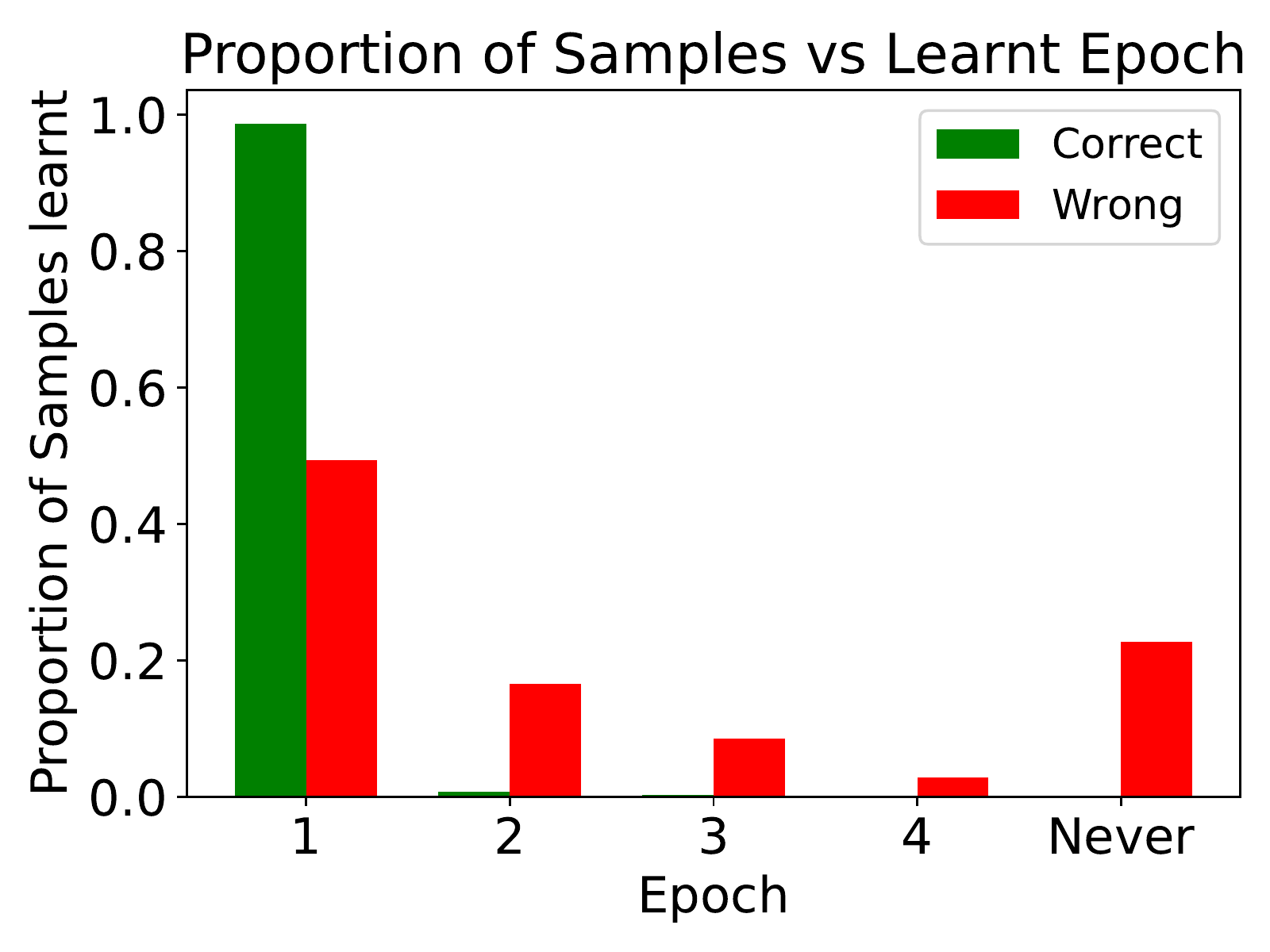}
        \label{figure:ep_filter}
    }
    \vspace{-3mm}
    \caption{
        Distributions of correct and wrong instances using different pseudo-label selection strategies on the NYT-Coarse dataset for its initial pseudo-labels.
        The base classifier is BERT.
        (a) is based on the softmax probability of samples' pseudo-labels and (b) is based on the earliest epochs at which samples are learnt.
    }
    \label{figure:filter}
    \vspace{-5mm}
\end{figure}

Weakly supervised text classification methods~\cite{agichtein2000snowball,riloff2003learning,tao2015doc2cube,meng2018weakly, mekala2020contextualized, mekala2020meta, mekala2021coarse2fine} typically start with generating pseudo-labels, and train a deep neural classifier to learn the mapping between documents and classes.
There is no doubt that the quality of pseudo-labels plays a fundamental role in the final classification accuracy, however, they are inevitably noisy due to their heuristic nature.
Pseudo-labels are typically generated by some heuristic, for example, through string matching between the documents and user-provided seed words~\cite{mekala2020contextualized}.
Deep neural networks (DNNs) trained on such noisy labels have a high risk of making erroneous predictions. 
More importantly, when self-training is employed, such error can be further amplified upon boostrapping.

To address this problem, in this paper, we study the pseudo-label selection in weakly supervised text classification, aiming to select a high quality subset of the pseudo-labeled documents (in every iteration when using self-training) that can potentially achieve a higher classification accuracy. 

\begin{figure*}[t]
    \center
    \includegraphics[width=\linewidth]{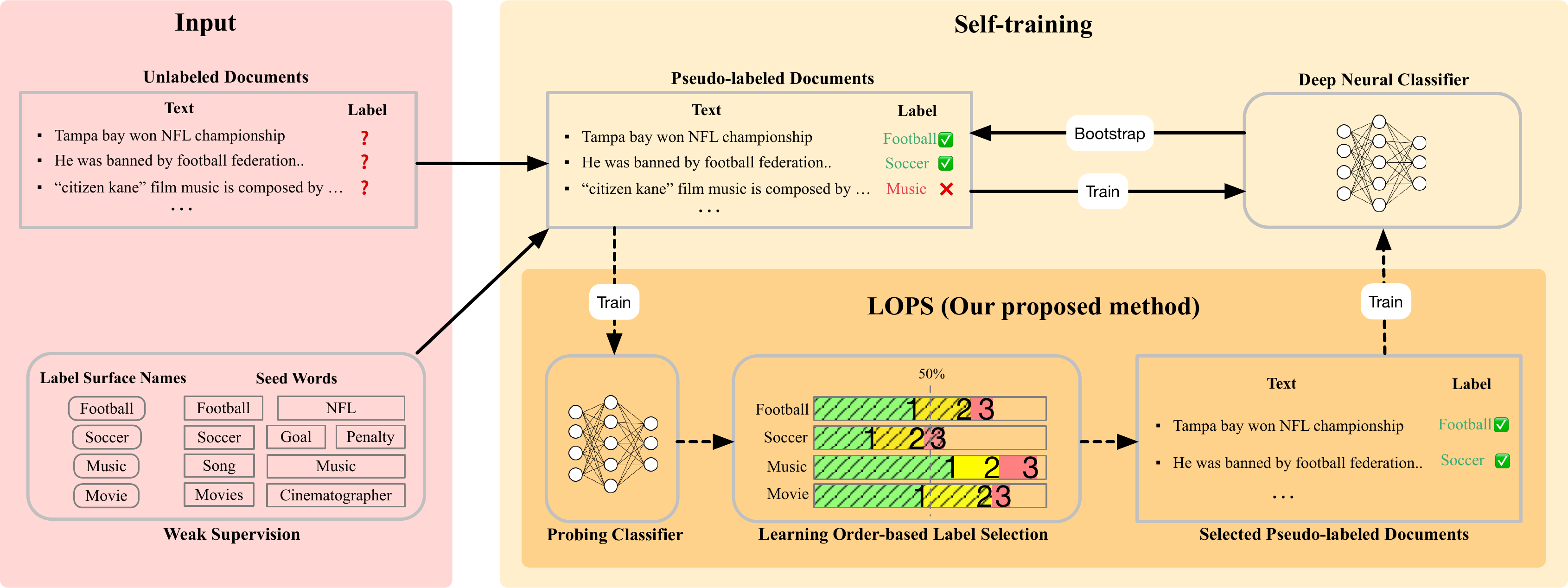}
    \caption{
    An overview of our proposed \our and how it plugs into self-training frameworks to replace the conventional training step. 
    Given pseudo-labeled samples, \our trains a probing classifier to obtain their learning order and we stop the training when at least $\tau \%$ of samples corresponding to each class are learnt and select the learnt samples. The numbers shown are learnt epochs and the samples in the shaded part are selected. A text classifier is trained on selected pseudo-labeled documents that is further used for inference and bootstrapping.
    }
    \label{fig:overview}
\end{figure*}

A straightforward solution is to first train a deep neural classifier based on the pseudo-labeled documents and then threshold the documents by the predicted probability scores corresponding to their pseudo-labels.
However, DNNs usually have a poor calibration and generate overconfident predicted probability scores~\cite{guo2017calibration}. 
For example, on New York Times (NYT) coarse-grained dataset, as shown in Figure~\ref{figure:prob_filter}, $60\%$ of wrong instances in the pseudo-labeled documents have a predicted probability by BERT greater than $0.9$ for their wrong pseudo-labels.



Recent studies on the memorization effects of DNNs show that they memorize easy and clean instances first, and gradually learn hard instances and eventually memorize the wrong annotations~\cite{arpit2017closer, geifman2018bias, zhang2021understanding}. 
We have confirmed this in our experiments for different classifiers. 
For example, as shown in Figure~\ref{figure:ep_filter}, BERT classifier learns most of the clean instances in the first epoch and learns wrong instances across all epochs.
Although it also learns good number of wrong instances in the first epoch, it is significantly less than the probability-based selection in Figure~\ref{figure:prob_filter}.
Therefore, we define the learning order of a pseudo-labeled document as the epoch when it is learnt during training i.e. when the training model's prediction is the same as its given pseudo-label.
Since correct samples are learnt first, we hypothesize that learning order-based selection will be able to filter out wrongly labeled samples.

Inspired by our observation, we propose a novel \underline{l}earning \underline{o}rder inspired \underline{p}seudo-label \underline{s}election method \our, as shown in Figure~\ref{fig:overview}.
Specifically, \our involves training a probing classifier on pseudo-labeled data and tracking the learning order of samples. 
We define a sample is learnt if and only if the classifier trained on pseudo-labels gives the same argmax prediction as its pseudo-label at the end of an epoch. 
We stop the training when at least $\tau \%$ of samples corresponding to each class are learnt and select all the learnt samples.
Then, we train a text classifier on these selected pseudo-labeled documents that is further used for inference.
We empirically show that \our can boost the accuracy of various weakly supervised text classification methods and it is much more effective and stable than probability score-based selections.

Our contributions are summarized as follows:
\begin{itemize}[leftmargin=*,nosep]
    \item We propose a novel pseudo-label selection method \our that takes learning order of samples into consideration.
    \item We show that selection based on learning order is much stable and effective than selection based on probability scores.
    \item Extensive experiments and case studies on real-world datasets with different classifiers and weakly supervised text classification methods demonstrate significant performance gains upon using \our.
    It can be viewed as a solid performance-boost plug-in for weak supervision.
\end{itemize}
\noindent\textbf{Reproducibility.} We will release the code and datasets on Github\footnote{\url{https://github.com/dheeraj7596/LOPS}}. 

\section{Related Work}
\label{rel-work}

\smallsection{Pseudo-Labels in Weakly Supervised Text Classification}
Since the weakly supervised text classification methods lack gold annotations, pseudo-labeling has been a common phenomenon to generate initial supervision. 
Pseudo-labeling depends on the type of weak supervision.
\citet{mekala2020contextualized} and \citet{mekala2020meta} have a few label-indicative seed words as supervision and they generate pseudo-labels using string-matching where a document is assigned a label whose aggregated term frequency of seed words is maximum.
\cite{meng2018weakly} generates pseudo-documents using the seed information corresponding to a label.
\cite{wang2020x} takes only label names as supervision and generates class-oriented document representations, and cluster them to create a pseudo-training set.
Under the same scenario, ~\cite{mekala2021coarse2fine} consider samples that exclusively contain the label surface name as its respective weak supervision.
In ~\cite{karamanolakis2021self-training}, pseudo-labels are created from the predictions of a trained neural network.
~\cite{arachie2021constrained} combines different weak signals to produce soft labels.

\smallsection{Label Selection}
There are different lines of work aiming to select true-labeled examples from a noisy training set.
One line of work involves training multiple networks to guide the learning process.
Along this direction, ~\cite{malach2017decoupling} maintains two DNNs and update them based on their disagreement.
~\cite{jiang2018mentornet} learns another neural network that provides data-driven curriculum.
~\cite{han2018co, yu2019does} use co-training where they select instances based on small loss criteria and cross-train two networks simultaneously.
~\cite{huang2019o2u} considers the training loss as the metric to filter out noise.
~\cite{swayamdipta-etal-2020-dataset} uses model's confidence and its variability across epochs to identify wrongly labeled samples.
Another line of work learns weights for the training data. 
Along this line, ~\cite{ren2018learning} propose a meta-learning algorithm that learns weights corresponding to training examples based on their gradient directions.
~\cite{fang2020rethinking} learns dynamic importance weighting that iterates between weight estimation and weighted classification. 
Recently, ~\cite{rizve2021defense} propose utilizing uncertainty to perform label selection.

\smallsection{Training dynamics}
In deep learning regime, models with large capacity are typically more robust to outliers. Nevertheless, data examples can still exhibit diverse levels of difficulties. \citet{arpit2017closer} finds that data examples are not learned equally when injecting noisy data into training. Easy examples are often learned first. \citet{Hacohen2019LetsAT} further shows such order of learning examples is shared by different random initializations and neural architectures. \citet{Toneva2019AnES} shows that certain examples are forgotten frequently during training, which means that they can be first classified correctly, then incorrectly. Model performance can be largely maintained when removing those least forgettable examples from training.

\begin{table}[t]
\centering
\caption{Noise ratios of different pseudo-label heuristics on NYT-Fine dataset. 
}
\label{weaksup_source}
\vspace{-3mm}
\small

\resizebox{\linewidth}{!}{
\begin{tabular}{lc}
\toprule
\textbf{Pseudo-label Heuristic} & \textbf{Noise Ratio} \\
\midrule
vMF distribution modeling~\cite{meng2018weakly} & 46.17\% \\
String-Match~\cite{mekala2020meta}  & 31.80\% \\
Contextualized String-Match~\cite{mekala2020contextualized}  & 31.24\% \\
Exclusive String-Match~\cite{mekala2021coarse2fine} & 52.13\% \\
Clustering~\cite{wang2020x} & 15.64\% \\
\bottomrule
\end{tabular}

}
\vspace{-3mm}

\end{table}

\section{Problem and Motivation}
\label{sec:formulation}

\noindent\textbf{Weakly supervised classification} refers to the problem with inputs (1) a set of unlabeled text documents $\mathcal{S} = \{x\}$, where $x \in \mathcal{X}$. 
(2) and $M$ target labels $\mathcal{C} = \{1, \ldots, M\}$. Our goal is to find a labeling function $f: \mathcal{X} \rightarrow \mathcal{C}$ that maps every document $x$ to its true label. Here we denote $y^*$ as the \emph{unknown} true label of a document $x$.
To cold start the classification of unlabeled documents, a source of weak supervision has to be introduced, which can come from various sources such as label surface names~\cite{wang2020x}, label-indicative seed words~\cite{mekala2020contextualized}, or rules~\cite{karamanolakis-etal-2021-self}. Given a ``weak'' labeling function $w: \mathcal{X} \rightarrow \mathcal{C}$, pseudo-labels are then generated on a subset of the unlabeled documents, which yields a labeled subset $\mathcal{D} = \{(x, w(x))\}$. For convenience, we denote $\mathcal{D}[j]$ to be the set of all documents that are pseudo-labeled as class $j$ in $\mathcal{D}$, namely $\mathcal{D}[j] = \{(x, w(x))\in \mathcal{D} | w(x) = j\}$.

\noindent\textbf{Pseudo-labels are noisy} due to their heuristic nature. 
For example,
on the NYT fine-grained dataset, we generate pseudo-labels using five different strategies~\cite{meng2018weakly, mekala2020contextualized, mekala2020meta, mekala2021coarse2fine, wang2020x} and compute their noise ratios.
As expected, no strategy is perfect and all of them generate noisy labels, ranging from 15\% to 50\% (see Table~\ref{weaksup_source}).

When a classifier is trained on such noisy training data, it can make some high confident erroneous predictions.
And, upon bootstrapping the classifier on unlabeled data, it has a snowball effect where such high confident erroneous predictions are added to the training data, and thus corrupting it more. 
As this process repeats for a few iterations, it adds more noise and significantly affects the final performance.
Therefore, identifying and selecting the correctly labeled samples is necessary 
and has a huge potential for a boost in performance. 
Note that, if the labels are not selected carefully, it could instead hurt the performance.

\smallsection{Our pseudo-label selection problem}
The weak supervision is likely to generate a noisy labeled set, which means $w(x) \ne y^*$ for some documents $x$. We denote $\mathcal{D}_\checkmark$ as the set of correctly labeled documents and $\mathcal{D}_\times = \mathcal{D} \setminus \mathcal{D}_\checkmark$ as the set of wrongly labeled documents, where $\mathcal{D}_\checkmark = \{ (x, w(x)) | w(x) = y^*\}$.
The problem of pseudo-label selection is thus to identify $\mathcal{D}_\checkmark$.

Note that pseudo-label selection is conceptually related to failure prediction~\cite{Hecker2018FailurePF, Jiang2018ToTO, Corbire2019AddressingFP} and out-of-distribution detection~\cite{Hendrycks2017ABF, Devries2018LearningCF, Liang2018EnhancingTR, Lee2018ASU}. However, the major difference here is for pseudo-label selection we have to detect wrong annotations in the training phase instead of inference phase.

\renewcommand\theadfont{}
\section{Our \our Framework}
In this section, we explain our framework \our in detail.
First, we 
give an overview of confidence function-based pseudo-label selection and discuss probability score as confidence function.
Then, we explain learning order as confidence function.
Finally, we show our algorithm that performs selection based on learning order.

\subsection{Overview: Confidence function-based Pseudo-label Selection}


In this section, we briefly introduce confidence function and discuss commonly-used probability score as confidence function. 

\noindent\textbf{Confidence function}
$\kappa: \mathcal{X}\times \mathcal{C}\rightarrow [0,1]$, assigns a value to each labeled document, which represents our confidence of its pseudo-label being correct.
Then, we can perform the selection by choosing a threshold $\gamma$ on confidence function.
We denote the set of labeled documents selected based on $\kappa$ and $\gamma$ as $\mathcal{\hat{D}}_\checkmark(\kappa, \gamma)$, namely
\begin{equation*}
    \mathcal{\hat{D}}_\checkmark(\kappa, \gamma) =  \{(x, w(x))\in \mathcal{D} ~|~ \kappa(x, w(x)) > \gamma\}
\end{equation*}
An optimal confidence function $\kappa^*$ should be able to perfectly distinguish the correctly labeled documents from wrongly labeled ones, namely there exists a threshold $\gamma^*$ such that
$
    \mathcal{\hat{D}}_\checkmark(\kappa^*, \gamma^*) = \mathcal{D}_\checkmark.
$
\smallsection{Probability score as confidence function}
One commonly-used intuitive confidence function for pseudo-label selection is the model's prediction probability scores corresponding to the pseudo-labels.
Probability scores have been used as confidence functions to select samples for bootstrapping~\cite{meng2018weakly, meng2019weakly, mekala2020contextualized}.
Specifically, let $\mathbf{f}: \mathcal{X} \rightarrow [0,1]^{|\mathcal{C}|}$ be a probabilistic classifier trained on pseudo-labeled documents and $\mathbf{f}(x)[j]$ represents the predicted probability of document $x$ belonging to class $j$, $\mathbf{f}(x)[w(x)]$ is used as the confidence function.
However, due to the poor calibration of DNNs~\cite{guo2017calibration}, probability scores of wrongly labeled documents are usually high.
As a result, it might be difficult to distinguish correctly- and wrongly-labeled documents based on probability scores.

\subsection{\our: Learning Order as Confidence Function}

\smallsection{Learning order}
Learning order of a pseudo-labeled document is the epoch when it is learnt during training, or more specifically when its label predicted by the model matches its given pseudo-label. 
Recent studies show that a DNN learns clean samples first and then gradually memorizes the noisy samples~\cite{arpit2017closer}.
We thus hypothesize that learning order can reflect the probability of wrong pseudo-label in terms of ranking.

We now utilize learning order to define a confidence function.
Specifically, let $\mathbf{f}^t(\cdot)$ be the classifier being trained at epoch $t$, and $T$ as the total number of epochs, the learning order of document $x$ can be defined as
\begin{equation}
\small
    \label{eq:learning-order}
    \begin{aligned}
        \eta(x, w(x)) = & 1 - \frac{1}{T}\min \{t ~|~ \arg\max_j \mathbf{f}^t(x)[j] = w(x)\},
    \end{aligned}
\end{equation}
where $t\in \{1,\ldots, T\}$. Here we have negated and scaled the learning order to be complied with the convention of confidence function i.e. higher confidence implies higher probability of a correct label.
We calculate the learning order at the granularity of epoch because the model would have seen all the training data by the end of an epoch, and hence, the learning order computed would be fair for all documents.
In case when the epoch number is not sufficient to distinguish the documents, one can increase the granularity of the learning order, for example, the batch number at which the document is learnt. Granularity higher than the epoch incurs extra training cost as a document will be examined more than once in each epoch.

\begin{algorithm}[t]
\caption{\our Method}\label{algorithm:1}
\SetAlgoLined
\small
  \textbf{Input:} A set of documents $\mathcal{D}$ pseudo-labeled by $w$, Probing Classifier $\mathbf{f}$.\\
  \textbf{Output:} Selected documents $\mathcal{\hat{D}}_{\checkmark}$ \\
  \For{{\bf epoch} $t = 1,2,\ldots, T$}{
    Train $\mathbf{f}$ on $\mathcal{D}$\\
    \For{$(x, w(x)) \in \mathcal{D}$} {
        \If {$\arg\max_j \mathbf{f}(x)[j] = w(x)$}{
            \If {$|\mathcal{\hat{D}}_{\checkmark}[w(x)]| / |\mathcal{D}[w(x)]| < \tau\%$} {
                $\mathcal{\hat{D}}_{\checkmark} = \mathcal{\hat{D}}_{\checkmark} \cup \{(x, w(x))\}$ \\
            }
        }
    }
    \If {$|\mathcal{\hat{D}}_{\checkmark}[j]| / |\mathcal{D}[j]| \ge \tau\%$ {\bf for all} $j$} {
        \textbf{Break} \\
    }
  }
  \textbf{Return} $\mathcal{\hat{D}}_{\checkmark}$\\
\end{algorithm}

\subsection{LOPS: Putting it all together}
Motivated by previous analyses, we utilize learning order to select pseudo-labels. 
We train a probing classifier on all pseudo-labeled documents and track their first learnt epoch during training. The confidence function can then be calculated based on Equation~(\ref{eq:learning-order}). 
Finally, we rank the documents based on their confidence and select the top-$\tau\%$ for each label independently.

To maximize the efficiency of LOPS, we utilize the fact that the top-ranked documents are learned earlier, and conduct the confidence calculation and pseudo-label selection simultaneously during training. 
Specifically, for each label, a document is selected once it is learnt, until the fraction of selected documents exceeds $\tau\%$ in this label. 
Whenever the fractions of selected documents exceeds $\tau\%$ for all labels, we stop the training. 
The pseudo-code is shown in Algorithm~\ref{algorithm:1}.
Note that \our can be plugged to any self-training based weakly-supervised classification framework as shown in Algorithm~\ref{algorithm:2}.


\begin{algorithm}[t]
\caption{Self-training with \our}\label{algorithm:2}
\SetAlgoLined
\small
  \textbf{Input:} Unlabeled data $\mathcal{D}$, Classifier $C$, Weak Supervision $w$.\\
  \textbf{Output:} Prediction labels $predLabs$ \\
  $\mathcal{\hat{D}}$ = Generate Pseudo-labels for $\mathcal{D}$, $w$\\
  \For{{\bf iteration} $it = 1,2,\ldots, n_{its}$}{
        $\mathcal{D}_{sel}$ = \our($\mathcal{\hat{D}}$, $C$)\\
        Train $C$ on $\mathcal{D}_{sel}$\\
        $predLabs$, $predProbs$ = Predict($C$, $\mathcal{D}$)\\
        $\mathcal{\hat{D}}$ = $\mathcal{\hat{D}}$ $\cup$ \{$x$ | $predProbs(x) > \delta$\}\\
    }
  \textbf{Return} $predLabs$\\
\end{algorithm}

\section{Experiments}
\label{sect:exp}

\begin{table}[t]
    \center
    \caption{Dataset statistics. 
    }
    \label{tbl:datastats}
    \vspace{-3mm}
    \small
    \scalebox{0.95}{
    \begin{tabular}{c c c c}
        \toprule
            {\textbf{Dataset}} & {\textbf{\# Docs}} & {\textbf{\# labels}} & {\textbf{Noise Ratio(\%)}} \\
        \midrule
        \textbf{NYT-Coarse} & 13,081 & 5 & 11.47 \\
        \textbf{NYT-Fine} & 13,081 & 26 & 31.80 \\
        \textbf{20News-Coarse} & 17,871 & 5 & 12.50\\
        \textbf{20News-Fine} & 17,871 & 17 & 25.67\\
        \textbf{AGNews} & 120,000 & 4 & 16.26\\
        \textbf{Books} & 33,594 & 8 & 37.32\\
        \bottomrule
    \end{tabular}
    }
\end{table}

\begin{table*}[t]
\centering
\caption{Evaluation results on six datasets using different combinations of classifiers and pseudo-label selection methods. 
Initial pseudo-labels are generated using String-Match. 
Micro- and Macro-F1 scores are used as evaluation metrics. Each experiment is repeated with three random seeds, mean and their respective standard deviations are presented in percentages.
For a fair comparison, we consider the same number of samples for all baselines as \our in each iteration.
Abnormally high standard deviations are highlighted in \textcolor{blue}{\textit{blue}} and low performances are highlighted in \textcolor{red}{\textit{red}}. \our outperforming \textit{Standard} is made bold and baselines performing better than our method are made bold. Statistical significance results are in Appendix~\ref{sec:statsig}.
}
\small
\resizebox{\linewidth}{!}{
\setlength{\tabcolsep}{1mm}
\begin{tabular}{c c cc cc cc cc cc cc}
\toprule
 & & \multicolumn{8}{c}{Coarse-grained Datasets} & \multicolumn{4}{c}{Fine-grained Datasets} \\
 \cmidrule(lr){3-10} \cmidrule(lr){11-14}
 & & \multicolumn{2}{c}{NYT-Coarse} & \multicolumn{2}{c}{20News-Coarse} & \multicolumn{2}{c}{AGNews} & \multicolumn{2}{c}{Books} & \multicolumn{2}{c}{NYT-Fine} & \multicolumn{2}{c}{20News-Fine}\\
 \cmidrule(lr){3-4} \cmidrule(lr){5-6} \cmidrule(lr){7-8} \cmidrule(lr){9-10} \cmidrule(lr){11-12} \cmidrule(lr){13-14}
Classifier & Method & Mi-F1 & Ma-F1 & Mi-F1 & Ma-F1 & Mi-F1 & Ma-F1 & Mi-F1 & Ma-F1 & Mi-F1 & Ma-F1 & Mi-F1 & Ma-F1 \\
\midrule
\multirow{9}{*}{BERT} & Standard & 90.1(0.17) & 80.3(0.91) & 77.3(0.27) & 76.4(0.76) & 75.4(0.64) & 75.4(0.47) & 55.7(0.54) & 57.9(0.82) & 77.2(0.36) & 71.6(0.43) & 70.0(0.30) & 69.6(0.25) \\
                      & \our & \textbf{94.6(0.36)} & \textbf{88.4(0.50)} & \textbf{81.7(1.00)} & \textbf{80.7(0.43)} & \textbf{79.5(0.86)} & \textbf{79.5(0.58)} & \textbf{57.7(0.87)} & \textbf{59.5(0.46)} & \textbf{84.3(0.54)} & \textbf{81.6(0.34)} & \textbf{73.8(0.61)} & \textbf{72.7(1.00)}\\
\cmidrule{2-14}
                        & MC-Dropout & 89.3(0.41) & 79.3(0.45) & 80.7(0.17) & 77.7(0.24) & 75.8(0.34) & 75.0(0.41) & 55.1(0.15) & 56.7(0.61) & 72.1(0.74) & 69.0(0.41) & 68.0(0.21) & 68.7(0.26)\\
                        & Entropy & 91.2(0.41) & 83.1(0.47) & 80.4(0.23) & 78.0(0.54) & \textbf{80.4(0.47)} & \textbf{80.0(0.42)}  & 55.2(0.74) & 56.7(0.42) & \textcolor{red}{43.4}(\textcolor{blue}{9.84})  & \textcolor{red}{18.1}(\textcolor{blue}{6.98}) & 64.3(0.74) & 63.6(0.83) \\
                        & O2U-Net & 92.9(0.41)& 85.9(0.69) & 80.9(0.28) & 78.5(0.19) & 79.8(0.47) & 79.8(0.53)  & 55.8(0.27) & 56.8(0.36) & \textcolor{red}{14.7}(\textcolor{blue}{10.24}) & \textcolor{red}{8.70}(\textcolor{blue}{7.31}) & 71.1(0.36) & 71.2(0.75) \\
                      & Random & 90.3(0.47) & 80.9(0.47) & 79.0(1.00) & 76.8(1.50) & 76.3(0.35) & 76.3(0.65) & 56.1(0.18) & 58.2(0.35) & 78.4(0.94) & 71.7(0.47) & 71.4(0.50) & 70.6(1.00)  \\
                      & Probability & 92.3(1.50) & 85.1(2.00) & 78.6(2.50) & 77.5(3.00) & 77.4(1.25) & 77.6(1.34) & 54.3(1.12) & 56.5(1.43) & \textcolor{red}{46.6}(2.50) & \textcolor{red}{22.3}(0.50) & \textcolor{red}{47.8}(\textcolor{blue}{23.50}) & \textcolor{red}{47.9}(\textcolor{blue}{23.50}) \\
                      & Stability & 93.3(0.50) & 86.5(0.50) & 76.7(5.00) & 75.4(5.00) & 79.3(0.75) & 79.5(0.35) & 55.0(0.43) & 57.0(0.19) & \textcolor{red}{48.1}(\textcolor{blue}{29.50}) & \textcolor{red}{35.5}(\textcolor{blue}{33.50})& 73.5(0.50) & 72.5(1.00) \\
\cmidrule{2-14}
                      & OptimalFilter & 98.3(0.27) & 96.4(0.37) & 94.7(0.37) & 94.9(0.61) & 89.4(0.46) & 89.3(0.76) & 76.2(0.21) & 76.7(0.19) & 97.4(0.71) & 92.2(0.62) & 87.6(0.37) & 86.5(0.36)\\
\midrule
\multirow{9}{*}{XLNet} & Standard & 89.2(0.74) & 80.1(0.64) & 77.6(0.39) & 75.4(0.68) & 72.7(0.97) & 72.4(0.53) & 57.6(0.31) & 58.7(0.46) & 77.4(0.34) & 71.3(0.75) & 60.7(0.74) & 66.5(0.61)\\
                    & \our & \textbf{89.5(0.17)} & \textbf{81.4(0.90)} & \textbf{82.5(0.50)} & \textbf{81.2(0.20)} & \textbf{77.7(0.57)} & \textbf{77.7(0.54)} & \textbf{58.5(0.65)} & \textbf{59.4(0.67)} & \textbf{80.7(0.22)} & \textbf{77.4(0.83)} & \textbf{70.6(0.31)} & \textbf{70.4(0.27)}\\
\cmidrule{2-14}
                        & MC-Dropout & 88.5(0.41) & 80.4(0.38) & 77.1(0.28) & 73.2(0.53) & 74.7(0.71) & 73.2(0.36) & 56.4(0.41) & 58.0(0.74) & 74.9(0.96) & 68.9(0.84) & 66.9(0.45) & 68.5(0.62) \\
                        & Entropy & \textbf{92.4(0.42)} & \textbf{85.4(0.51)} & 78.2(0.36) & 74.4(0.45) & 72.9(0.67) & 72.0(0.51) & 54.5(0.74) & 56.6(0.65) & 77.9(0.67) & 70.7(0.38) & 68.8(0.65) & 69.5(0.74) \\
                        & O2U-Net & 92.2(0.37) & 84.6(0.24) & 80.5(0.93) & 77.4(0.57) & 71.6(0.69) & 68.8(0.61) & 58.1(0.17) & \textbf{59.9(0.52)} & 79.6(0.47) & 76.8(0.59)  & 67.2(0.64) & 69.0(0.26) \\
                      & Random & 90.7(0.03) & 80.5(0.51) & 78.6(0.50) & 75.4(1.00) & 67.5(0.22) & 67.4(0.63) & 57.5(0.43) & 58.3(0.45) & 76.6(0.94) & 72.7(0.70) & 67.3(0.49) & 67.2(0.32)\\
                      & Probability & 91.3(0.29) & 83.4(0.50) & 77.4(1.00) & 75.2(0.30) & 70.1(1.09) & 70.4(1.14) & 54.6(1.42) & 56.3(1.26) & \textcolor{red}{38.2}(\textcolor{blue}{6.50}) & \textcolor{red}{36.5}(1.00) & 69.5(0.82) & 69.2(0.12)\\
                      & Stability & 91.4(1.00) & 82.3(1.50) & 79.7(1.50) & 77.6(1.50) & 74.3(1.10) & 74.5(0.87) & 56.3(0.88) & 58.1(0.97) & 79.5(0.50) & 76.3(1.10) & 68.5(0.49) & 68.4(1.00)\\
\cmidrule{2-14}
                      & OptimalFilter & 98.3(0.12) & 96.5(0.21) & 94.5(0.23) & 94.4(0.29) & 89.3(0.28) & 89.7(0.39) & 76.4(0.44) & 76.3(0.43) & 97.4(0.32) & 93.6(0.38) & 86.6(0.43) & 86.4(0.35) \\
\midrule
\multirow{9}{*}{GPT-2} & Standard & 91.1(0.24) & 82.3(0.28) & 78.4(0.26) & 76.3(0.38) & 61.3(0.28) & 61.2(0.43) & 51.6(0.41) & 53.3(0.37) & 76.2(0.41) & 69.5(0.38) & 70.5(0.46) & 70.4(0.38)\\
                        & \our & \textbf{95.2(0.49)} & \textbf{89.1(0.51)} & \textbf{82.5(0.57)} & \textbf{80.3(0.63)} & \textbf{75.7(0.52)} & \textbf{75.3(0.31)} & \textbf{56.8(0.89)} & \textbf{58.6(0.63)} & \textbf{80.4(0.09)} & \textbf{76.3(0.21)} & \textbf{70.6(0.76)} & \textbf{70.5(0.48)}\\
\cmidrule{2-14}
                        & MC-Dropout & 89.2(0.14) & 79.8(0.51) & 80.2(0.63) & 77.1(0.57) & 65.5(0.34) & 65.1(0.94) & 49.5(0.74) & 51.5(0.54) & 74.1(0.62) & 68.2(0.21) & 70.4(0.47) & 70.8(0.65)\\
                        & Entropy & 93.1(0.32) & 85.9(0.36) & 80.8(0.65) & 77.9(0.84) & 65.4(0.85) & 65.3(0.54) & 54.3(0.32) & 55.5(0.47) & 77.4(0.42) & 75.3(0.65) & 69.1(0.62) & 69.6(0.21) \\
                        & O2U-Net & 93.8(0.89) & 87.5(0.24) & 81.2(0.76) & 77.9(0.37) & 72.0(0.38) & 70.7(0.75) & 55.1(0.27) & 57.2(0.67)  & 80.2(0.41) & \textbf{79.4(0.58)}  & 70.3(0.24) & \textbf{71.4(0.16)} \\
                      & Random & 90.2(0.42) & 80.2(0.56) & 79.7(0.46) & 78.4(0.32) & 68.2(0.18) & 68.1(0.19) & 53.4(0.46) & 55.3(0.42) & 77.5(0.52) & 70.4(1.02) & 69.4(0.21) & 69.3(0.29)\\
                      & Probability & 93.3(1.04) & 85.5(1.13) & 80.4(1.49) & 78.5(1.50) & 66.2(0.69) & 66.6(0.89) & 51.7(1.11) & 54.5(1.09) & 76.7(0.57) & 71.3(0.69) & 69.4(1.21) & 69.3(1.18) \\
                      & Stability & 94.4(0.56) & 88.6(0.59) & 81.4(1.02) & 78.6(1.50) & 72.4(0.58) & 72.3(0.53) & 53.6(1.02) & 55.3(1.13) & 79.4(0.62) & 75.3(0.65) & \textbf{70.6(0.68)} & 70.4(0.63)\\
\cmidrule{2-14}
                      & OptimalFilter & 98.3(0.24) & 96.2(0.21) & 94.2(0.23) & 93.3(0.27) & 88.7(0.26) & 88.4(0.28) & 72.3(0.19) & 73.7(0.22) & 97.3(0.18) & 92.4(0.19) & 86.1(0.35) & 85.5(0.38)\\
\bottomrule
\end{tabular}
}
\label{results-cls}
\end{table*}

We evaluate our label selection method based on end-to-end classification performance using different state-of-the-art classifiers and  weakly supervised text classification frameworks. 
And also, we evaluate learning order as a confidence function and provide a comparison with probability score as confidence function.

\subsection{Datasets}


We experiment on four datasets: New York Times (\textbf{NYT}), 20 Newsgroups (\textbf{20News}), \textbf{AGNews}~\cite{zhang2015character}, \textbf{Books}~\cite{DBLP:conf/recsys/WanM18, DBLP:conf/acl/WanMNM19}. NYT and 20News datasets also have fine-grained labels which are also used for evaluation. 
Initial pseudo-labels are generated using String-Match~\cite{mekala2020contextualized}.
The dataset statistics and corresponding noise ratios of initial pseudo-labels are provided in Table~\ref{tbl:datastats} and more details are provided in Appendix~\ref{sec:datasets}.


\subsection{Compared Methods}
We compare with several label selection methods mentioned below:
\begin{itemize}[leftmargin=*,nosep]
    \item \textbf{O2U-Net:}~\cite{huang2019o2u} trains a classifier cyclically to make its status transfer from overfitting to underfitting and records losses of each sample. They consider the normalized loss as the metric to filter out the noise.
    \item \textbf{MC-Dropout:}~\cite{mukherjee2020uncertainty} performs pseudo-label selection based on uncertainty estimates computed using probability scores.
    \item \textbf{Entropy:} is similar to MC-Dropout, however uses entropy to compute uncertainty scores. 
    \item \textbf{Probability:} We sort the prediction probabilities corresponding to pseudo-labels in descending order and select the same number of samples as \our in each iteration of bootstrapping.
    \item \textbf{Random:} We randomly select the same number of samples as \our in each iteration of bootstrapping. To avoid skewed selection, we sample in a stratified fashion based on class labels.
    \item \textbf{Learning Stability (stability):} ~\cite{Dong2021DataQM} introduced a metric to measure the data quality based on the frequency of events that an example is predicted correctly throughout the training. We sort the samples based on learning stability in descending order i.e. most stable to least stable and select the same number of samples as \our in each iteration of bootstrapping.
\end{itemize}

\vspace{1mm}
\noindent To perform controlled experiments with a fair comparison, we consider the same number of samples as \our in each iteration for all above baselines because we cannot tune individual thresholds for each dataset since there is no human-annotated data under the weakly supervised setting and one fixed threshold for all datasets doesn't work as distribution of prediction probability varies across datasets.

We also present experimental results without any label selection in addition to the probability threshold $\delta$ while bootstrapping (denoted by \textbf{\textit{Standard}}) as lower bound and with all the wrongly annotated samples removed as \textbf{\textit{OptimalFilter}}.

\begin{table}[t]
\caption{Evaluation results of weakly supervised text classification frameworks with \our. 
This demonstrates that \our can be easily plugged in and improves the performance.
}
\centering
\small
\setlength{\tabcolsep}{0.5mm}
\resizebox{\linewidth}{!}{
\begin{tabular}{c cc cc cc cc cc cc}
\toprule
& \multicolumn{2}{c}{NYT-Coarse} & \multicolumn{2}{c}{NYT-Fine} & \multicolumn{2}{c}{20News-Coarse} & \multicolumn{2}{c}{20News-Fine} & \multicolumn{2}{c}{AGNews} & \multicolumn{2}{c}{Books}\\
 \cmidrule(lr){2-3} \cmidrule(lr){4-5} \cmidrule(lr){6-7} \cmidrule(lr){8-9} \cmidrule(lr){10-11} \cmidrule(lr){12-13}
Method & Mi-F1 & Ma-F1 & Mi-F1 & Ma-F1 & Mi-F1 & Ma-F1 & Mi-F1 & Ma-F1 & Mi-F1 & Ma-F1 & Mi-F1 & Ma-F1\\
\midrule
\multicolumn{13}{c}{ConWea} \\
\cmidrule(lr){2-13}
Standard & 93.1 & 87.2 & 87.4 & 77.4 & 74.3 & 74.6 & 68.7 & 68.7 & 73.4 & 73.4 & 52.3 & 52.6 \\
\our & \textbf{94.2} & \textbf{90.1} & \textbf{87.5} & \textbf{78.6}  & \textbf{79.7} & \textbf{78.4} & \textbf{70.4} & \textbf{70.6} & \textbf{79.2} & \textbf{79.2} & \textbf{57.5} & \textbf{58.7}  \\
\midrule
\multicolumn{13}{c}{X-Class} \\
\cmidrule(lr){2-13}
Standard & \textbf{96.3} & \textbf{93.3} & 86.6 & \textbf{74.7} & 58.2 & 61.1 & 70.4 & 70.4  & 82.4 & 82.3 & 53.6 & 54.2  \\
\our & 96.2 & \textbf{93.3} & \textbf{86.8} & 73.8  & \textbf{60.7} & \textbf{62.3} & \textbf{71.2} & \textbf{71.2} & \textbf{83.6} & \textbf{82.7} & \textbf{54.2} & \textbf{56.3} \\
\midrule
\multicolumn{13}{c}{WeSTClass} \\
\cmidrule(lr){2-13}
Standard & 92.3 & 86.0 & 67.1  & 60.4  & 53.2 & 49.4 & 54.9 &  54.9 & 80.4 & 80.1 & 49.7  & 48.1  \\
\our & \textbf{93.4} & \textbf{88.1} & \textbf{68.4}  & \textbf{63.8}  & \textbf{53.3} & \textbf{51.5} & \textbf{61.1} & \textbf{60.5} & \textbf{81.4} & \textbf{81.3} & \textbf{51.2}  & \textbf{49.8} \\
\midrule
\multicolumn{13}{c}{LOTClass} \\
\cmidrule(lr){2-13}
Standard & \textbf{70.1} & \textbf{30.3} & \textbf{5.3} & \textbf{4.1}  & \textbf{47.0} & \textbf{35.0} & \textbf{12.3} & \textbf{10.6} & 84.9 & 84.7 & \textbf{19.9}  & \textbf{16.1} \\
\our & \textbf{70.1} & \textbf{30.3} & 3.5 & 2.9  & 45.7 & 32.6 & 7.8 & 4.1 &  \textbf{86.2} & \textbf{86.1} & 15.8  & 10.3 \\
\bottomrule
\end{tabular}
}
\label{results-weaksup}
\end{table}

\subsection{Experimental Settings}

For all our experiments, we consider seed words used in ~\cite{mekala2020contextualized, wang2020x} as weak supervision and generate initial pseudo-labels using String-Match~\cite{mekala2020meta} unless specified. The average number of seeds are 4 per class.
We experiment on three state-of-the-art text classifiers: 
(1) \textbf{BERT} (\verb+bert-base-uncased+)~\cite{devlin2018bert},
(2) \textbf{XLNet} (\verb+xlnet-base-cased+)~\cite{yang2019xlnet},
and (3) \textbf{GPT-2}~\cite{radford2019language}.
We follow the same self-training method for all classifiers that starts with generating pseudo-labels, training a classifier on pseudo-labeled data, and bootstrap it on unlabelled data by adding samples whose prediction probabilities are greater than $\delta$. 
Following~\cite{mekala2020contextualized}, we assume that weak supervision $\mathcal{W}$ is of reasonable quality i.e. majority of pseudo-labels are good. Therefore, we set $\tau$ to $50\%$. While training the classifiers, we fine-tune BERT, XLNet, GPT-2 for $4$ epochs.
We bootstrap all the classifiers for $5$ iterations with the probability threshold $\delta$ as $0.6$. 
We also experiment on state-of-the-art weakly supervised text classification methods: \textbf{ConWea}~\cite{mekala2020contextualized}, \textbf{X-Class}~\cite{wang2020x}, \textbf{WeSTClass}~\cite{meng2018weakly}, and \textbf{LOTClass}~\cite{meng2020text}.
Three of them are self-training-based methods and more details are mentioned in Appendix~\ref{sec:methods}.

\subsection{End-to-End Classification Performance}




\subsubsection{Results: Different Classifiers}
We summarize the evaluation results with different combinations of classifiers and selection methods in Table~\ref{results-cls}.  
All experiments are run on three random seeds and mean, standard deviations are reported.


As shown in Table~\ref{results-cls}, upon plugging our proposed method \our, we observe a significant boost in performance consistently over \textbf{\textit{Standard}} with all the classifiers.
We observe that \our always outperforms random selection which shows that the selection in \our is strategic and principled.
\our performs better than probability and stability based selection methods in most of the cases. This shows that \our is very effective in removing wrongly labeled and preserving correctly labeled samples.
\our also performs better than O2U-Net~\cite{huang2019o2u} and MC-Dropout~\cite{mukherjee2020uncertainty} in most of the datasets demonstrating the effectiveness of learning order as confidence function.

We also observe a significant boost in performance over \textbf{\textit{Standard}} with all the classifiers in the case of fine-grained datasets as well.
In some cases like BERT on NYT-Fine, the improvement is as high as 7 points on micro-f1 and 10 points on macro-f1.
We observe abnormally low performances of probability and stability based selection methods in some scenarios (highlighted in \textcolor{red}{\textit{red}}).
This is because the number of noisy labels are more in fine-grained datasets and gets amplified with self-training and resulting in high noise.
Moreover, we also observe that probability and stability based selections are biased towards majority labels and select wrong majority labels over correct minority labels.
For example, the precision of pseudo-labels belonging to minority classes like \textit{cosmos}, \textit{gun control}, and \textit{abortion} in NYT-Fine before selection is 100\% and it selected almost none of these whereas it selected 700 wrong documents belonging to a majority labels like, \textit{international business}.
Although stratified selection can be employed to address this problem, this ends up having a same threshold and selecting a fixed ratio of samples for every dataset, which might not be optimal for every dataset.

We have to note unusually high standard deviation for probability and stability in some cases~(highlighted in \textcolor{blue}{\textit{blue}}).
This demonstrates that these selection methods are unstable. \our is comparatively more stable and its effectiveness is largely due to its invariance. Although these methods outperform \our in a few cases, their unstable nature makes them unreliable. 
Therefore, we believe \our is superior than compared methods.

\subsubsection{Results: Different Weakly-Supervised Text Classification Methods}
We summarize the evaluation results with different weakly supervised methods in Table~\ref{results-weaksup}.
The results demonstrate that \our improves the performance of ConWea and WeSTClass significantly on all datasets and X-Class sometimes. 
Note that, X-Class sets a confidence threshold and selects only top-50\% instances, which provides a hidden advantage and \our improves the performance on top of it for some datasets.
We have to note the significantly low performance of LOTClass. 
It is observed that LOTClass requires a wide variety of contexts of label surface names from the input corpus to generate high quality category vocabulary, which plays a key role in performance~\cite{wang2020x}.
The performance is comparitively worse in fine-grained classes than coarse-grained classes because LOTClass assumes that the replacements of label surface names are indicative of its respective label. However, this might not be a valid assumption for fine-grained classes~\cite{mekala2021coarse2fine}. 
Among the datasets we experimented on, these requirements are satisfied only by AGNews dataset where there are many documents(120000) classified broadly into 4 categories and we observe a performance boost using LOPS on this dataset. 
Due to poor quality of pseudo-labels for other datasets, there is no increment in performance with LOPS.

\begin{figure}[t]
    \centering
    \includegraphics[width=1.0\linewidth]{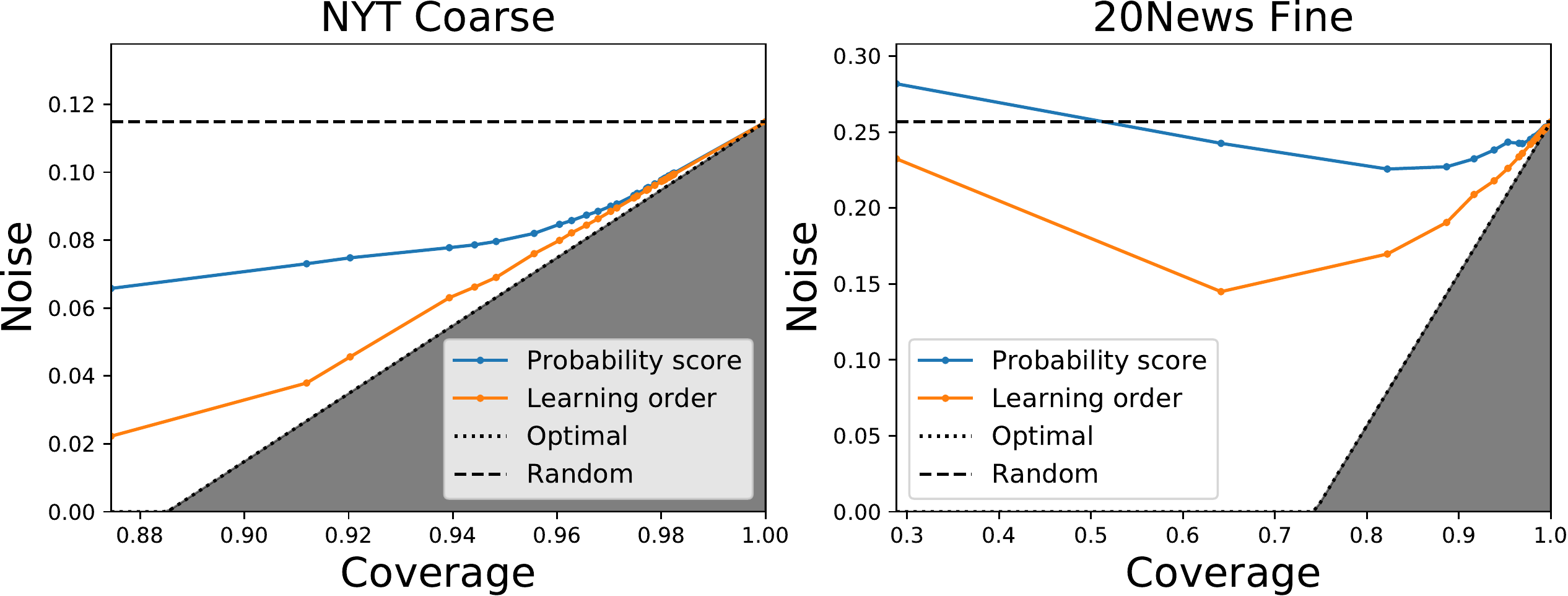}
    \vspace{-3mm}
    \caption{NC-curves of learning order and probability score with BERT as the classifier.}
    \label{figure:nc-curve}
    \vspace{-5mm}
\end{figure}

\begin{figure}[t]
    \subfigure[NYT Coarse]{
        \includegraphics[width=0.47\linewidth]{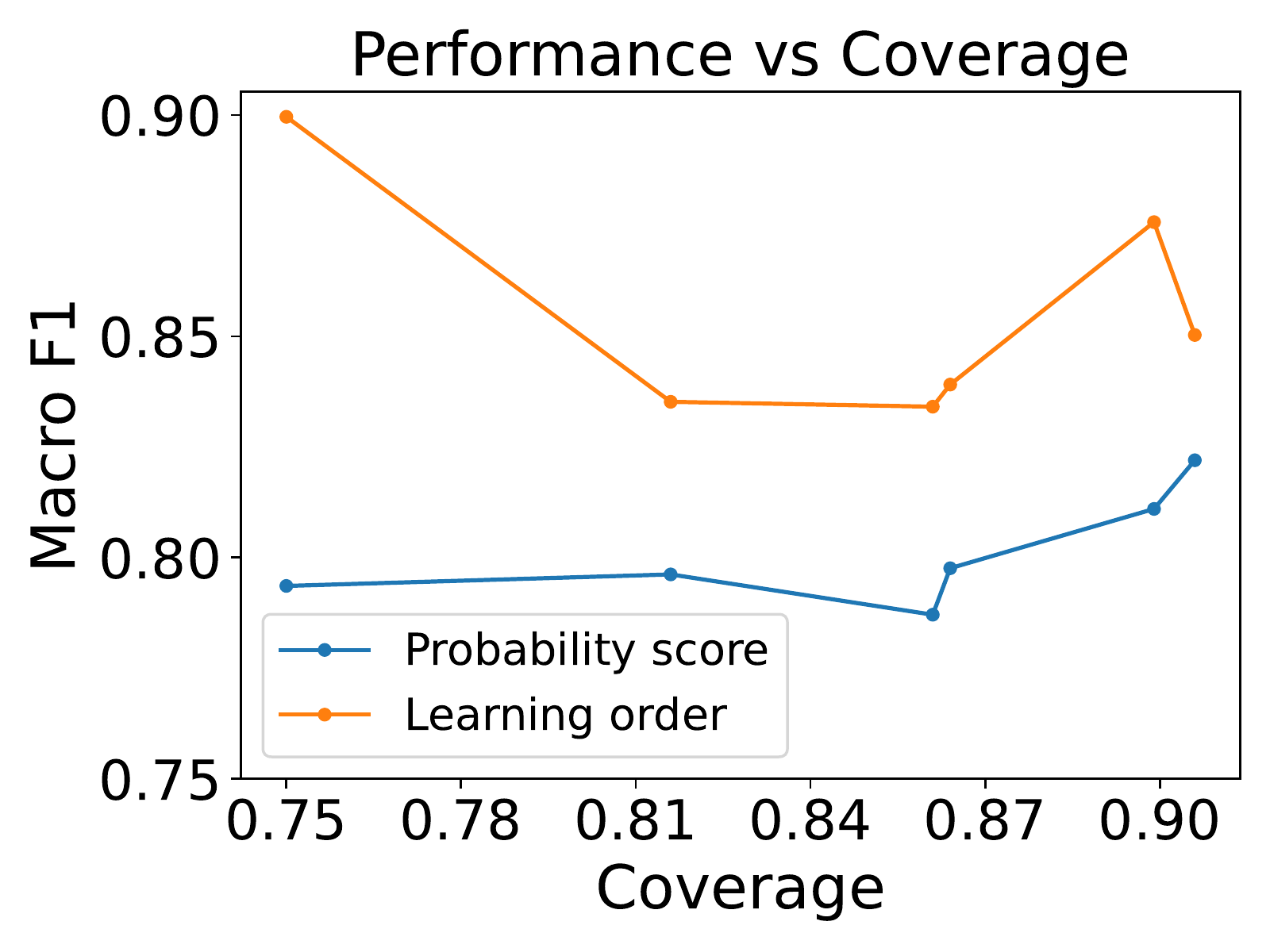}
    }
    \subfigure[20News Fine]{
        \includegraphics[width=0.47\linewidth]{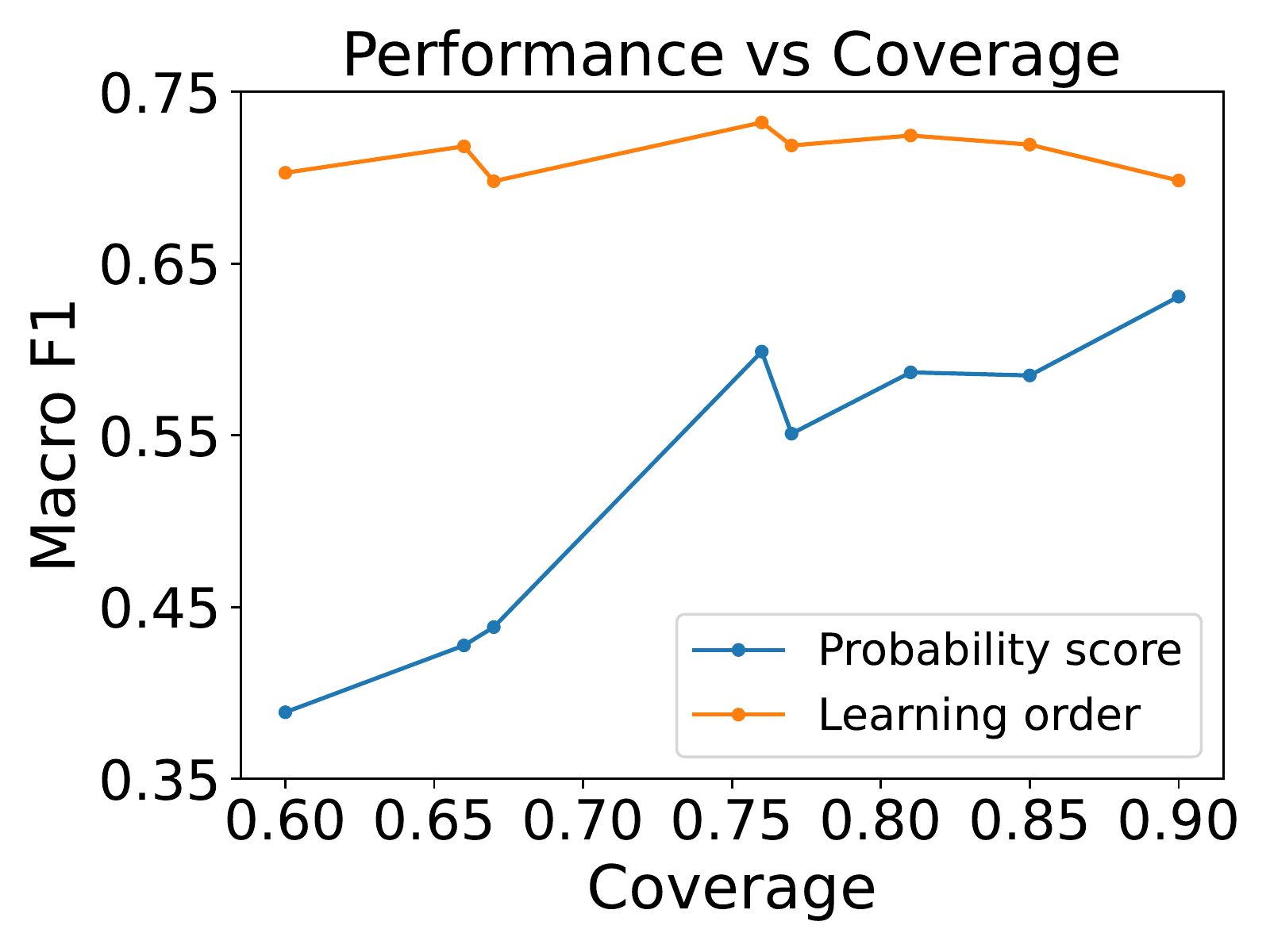}
    }
    \vspace{-3mm}
    \caption{Macro-F$_1$ vs Coverage on NYT-Coarse \& 20News-Fine using BERT with \our and Probability score based selection.}
    \label{figure:perf_coverage}
    \vspace{-5mm}
\end{figure}

\subsection{Learning Order vs Probability Score: Evaluating Confidence Functions}


In this section, we define evaluating a confidence function and compare learning order and probability score as confidence functions.


\smallsection{Evaluation of a confidence function}
Ideally, there exists a threshold for a given confidence function that perfectly distinguishes correctly and wrongly labeled samples.
However, in practice, confidence functions may not suffice such ideal condition.
There always exists a trade-off between \emph{noise} $\epsilon(\kappa, \gamma)$ and \emph{coverage} $\phi(\kappa, \gamma)$, defined as:
\begin{equation*}
\small
    \begin{aligned}
    \epsilon(\kappa, \gamma) & = \frac{|\mathcal{\hat{D}}_\checkmark(\kappa, \gamma) \cap \mathcal{D}_\times|}{|\mathcal{\hat{D}}_\checkmark(\kappa, \gamma)|},~~
    \phi(\kappa, \gamma) & = \frac{|\mathcal{\hat{D}}_\checkmark(\kappa, \gamma)|}{|\mathcal{D}|}.
    \end{aligned}
\end{equation*}
The coverage is the fraction of labeled documents being selected and the noise is the fraction of wrongly labeled documents within selected documents.
A small threshold leads to high coverage i.e. most labeled documents will be selected, thus being more noisy.
And a high threshold leads to an opposite situation.
Therefore, to evaluate a confidence function, we plot noise and coverage at various thresholds, which we refer as the \emph{noise-coverage curve} (NC-curve) and compute the \emph{area under the noise-coverage curve} (AUNC).
As shown in figure~\ref{figure:nc-curve}, an optimal confidence function selects wrongly labeled documents only after selecting all the correctly labeled documents, hence generates a NC-curve in the shape of a rectifier, namely $\epsilon = \max(0, \phi - |\mathcal{D}_\checkmark| / |\mathcal{D}|)$.
A random confidence function always selects the same fraction of wrongly labeled documents, hence an NC-curve with a constant value.
An ideal confidence function should minimize AUNC.


    



\smallsection{Learning Order vs Probability Score}
We plot NC-curves of learning order and probability scores in Figure~\ref{figure:nc-curve} with BERT classifier on NYT-Coarse, 20News-Fine datasets.
To isolate them from the effects of bootstrapping, we don't perform any bootstrapping.
We also plot the end-to-end performance vs coverage in Figure~\ref{figure:perf_coverage}.
From Figure~\ref{figure:nc-curve}, we observe that learning order has significantly smaller AUNC compared to the probability score.
In some datasets such as NYT-Coarse, it even approaches optimal confidence function.
In fine-grained datasets like 20News-Fine, the calibration is so poor that the probability score is even worse than random, which explains poor empirical results of Probability-based selection on fine-grained datasets.
From Figure~\ref{figure:perf_coverage}, we observe that the performance with LOPS is significantly better and more stable than Probability.

\begin{figure}[t]
    \subfigure[20News Coarse]{
        \includegraphics[width=0.47\linewidth]{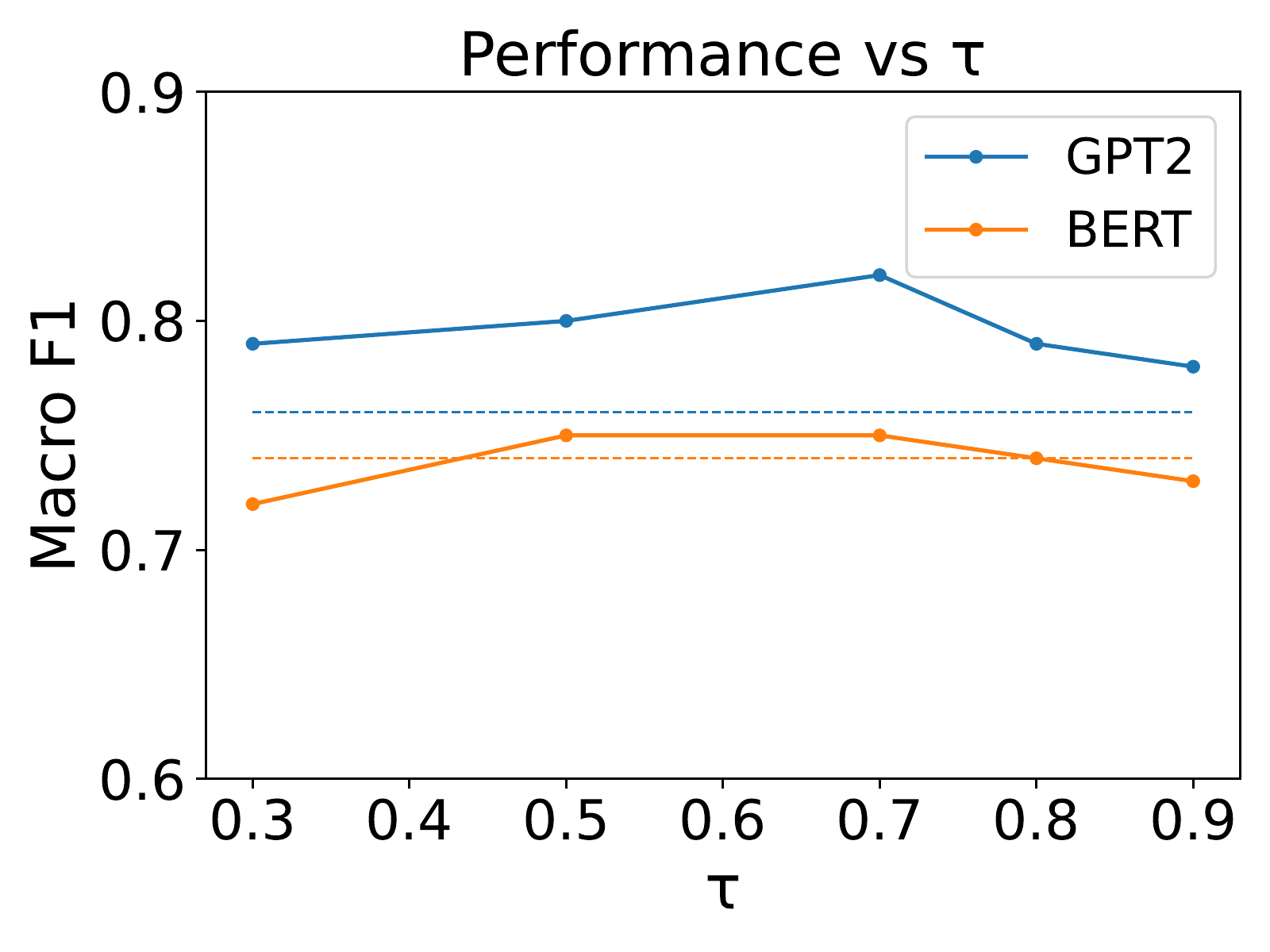}
    }
    \subfigure[Books]{
        \includegraphics[width=0.47\linewidth]{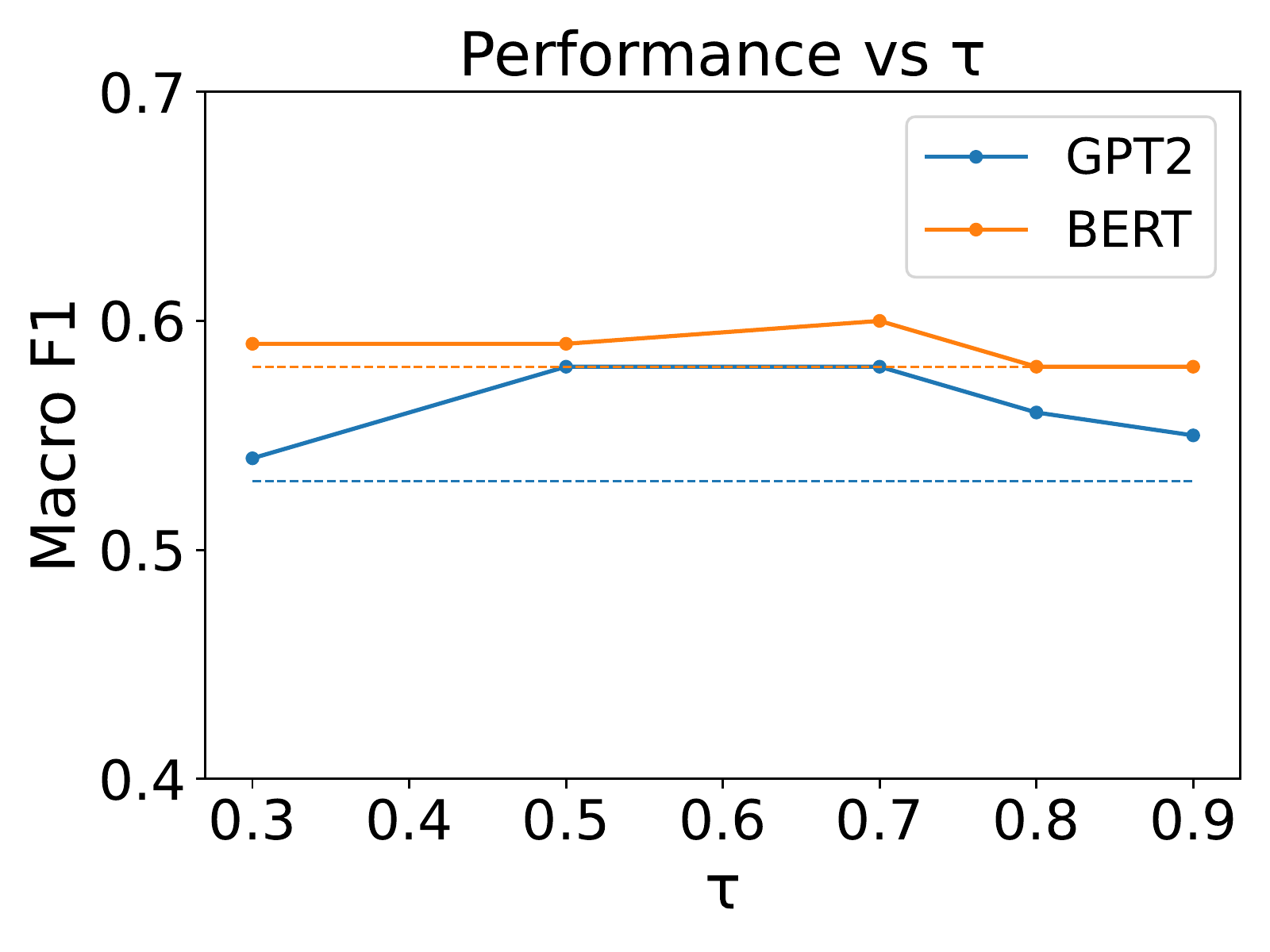}
    }
    \vspace{-3mm}
    \caption{Macro-F$_1$ vs $\tau$ on 20News-Coarse \& Books using GPT2 and BERT with \our. The dashed lines represent performance with no label selection.}
    \label{figure:tau}
    \vspace{-5mm}
\end{figure}

\begin{figure}[t]
    \subfigure[20News Fine]{
        \includegraphics[width=0.47\linewidth]{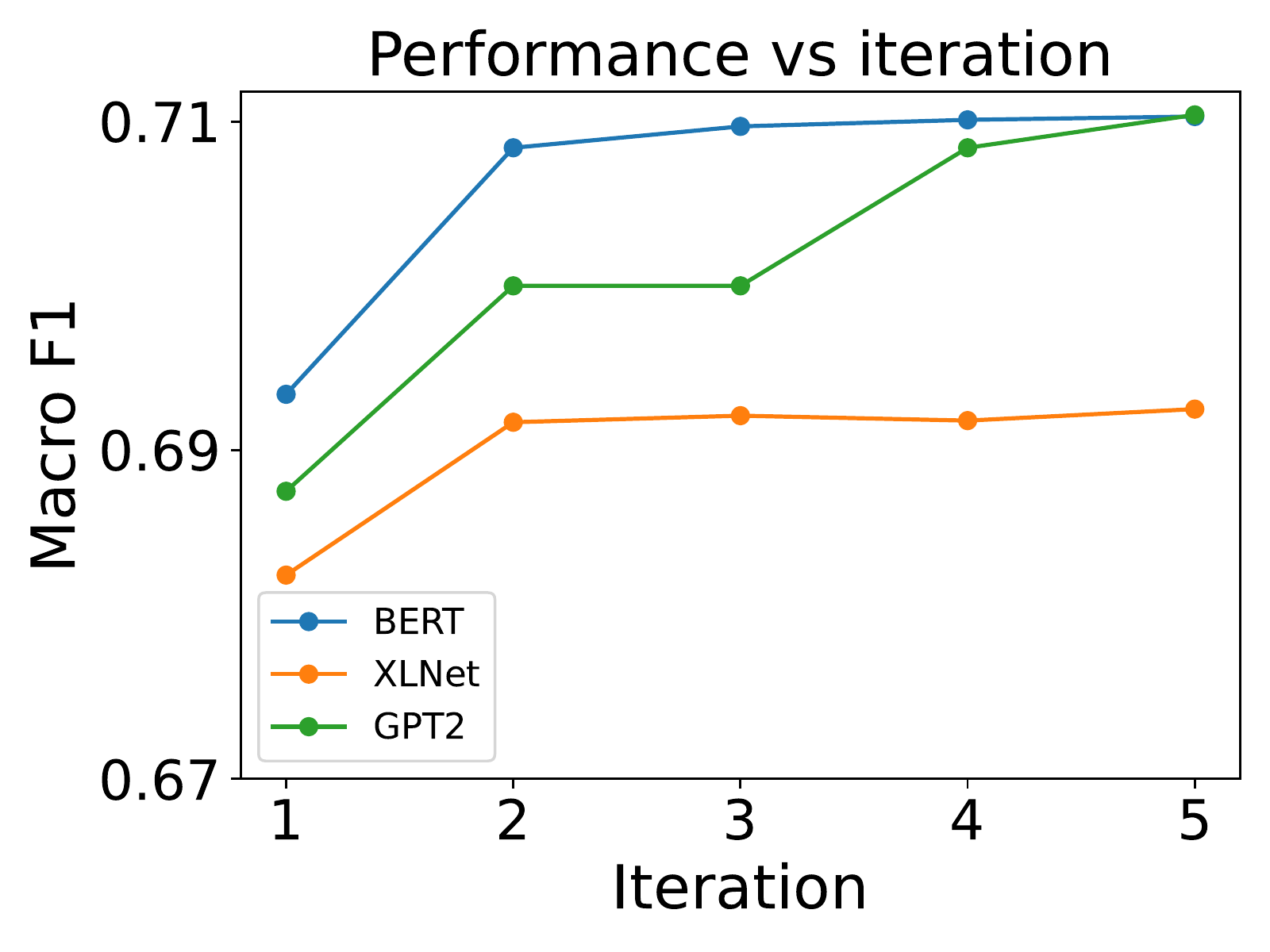}
    }
    \subfigure[Books]{
        \includegraphics[width=0.47\linewidth]{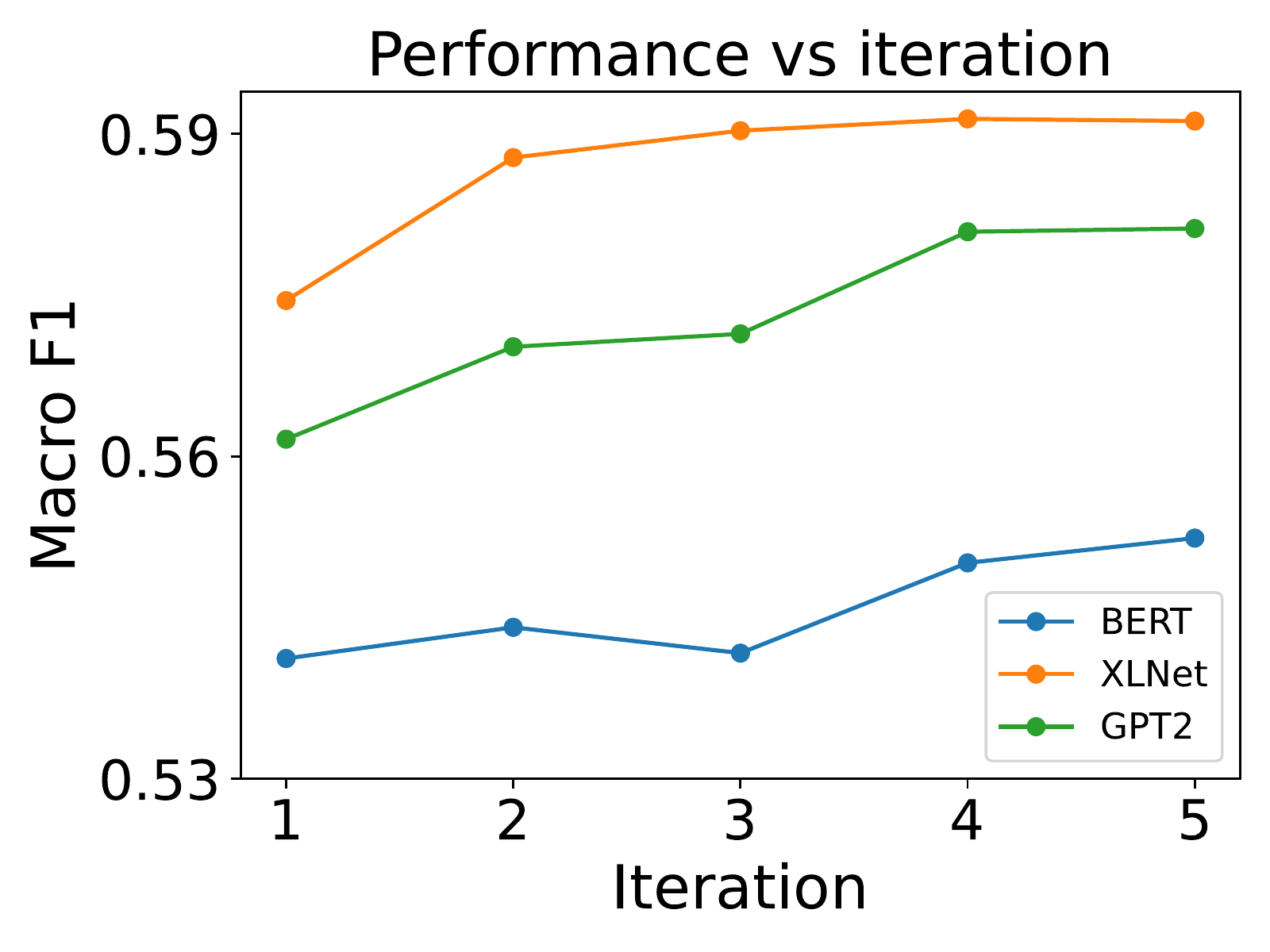}
    }
    \vspace{-3mm}
    \caption{Macro-F$_1$ vs iteration on 20News-Fine \& Books using BERT, XLNet, GPT2 with \our.}
    \label{figure:it}
    \vspace{-5mm}
\end{figure}

\subsection{Performance vs $\tau$}
To study the effect of $\tau$ on performance, 
we plot macro-f1 vs $\tau$ on 20News-coarse and Books datasets using GPT2 and BERT classifiers, shown in Figure~\ref{figure:tau}. 
We observe that the performance increases initially and gradually drops down at higher $\tau$ values.
The lower $\tau$ values imply being highly selective and thus the few number of selected samples are not enough for the model to generalize.
The higher $\tau$ values imply poor selection with many noisy labels, making the performance to drop. 
From the plot, we can observe that the performance is robust for middle $\tau$ values i.e. $50-70\%$. 

\subsection{Performance vs iteration}
The plot of performance vs the number of iteration of bootstrapping is shown in Figure~\ref{figure:it}. We observe that the macro f1 increases initially and gradually converges at the later iterations.


\section{Conclusion and Future Work}
\label{sect:con}
In this paper, we proposed \our, a novel learning order inspired pseudo-label selection method. 
Our method is inspired from recent studies on memorization effects that showed that clean samples are learnt first and then wrong samples are memorized.
Experimental results demonstrate that our method is effective, stable and can act as a performance boost plugin on many text classifiers and weakly supervised text classification methods.
In the future, we are interested in automatically identifying the right granularity to measure learning order for a given dataset.
Moreover, we are also interested in analyzing the learning order in classification tasks in image and speech domains.

\section{Limitations}

Since we select 50\% of the samples based on learning order, our method requires the absolute number of pseudo-labeled samples to be high enough so that the final classifier has significant number of selected samples to learn and generalize on. 
For example, we experimented on a subset of 2613 samples from 20news-fine dataset with noise rate 20\%. 
With \our, the macro f1 is 68.3\% and without any selection the macro-f1 is 70.1\%.
We attribute this performance drop to the lack of generalization using the few selected samples from \our.
Since in real-life scenario, obtaining noisy annotations is cheaper, we believe this limitation can be addressed comfortably.
\section{Acknowledgements}

We thank anonymous reviewers and program chairs for their valuable and insightful feedback. 
The research was sponsored in part by National Science Foundation Convergence Accelerator under award OIA-2040727 as well as generous gifts from Google, Adobe, and Teradata.
Any opinions, findings, and conclusions or recommendations expressed herein are those of the authors and should not be interpreted as necessarily representing the views, either expressed or implied, of the U.S. Government. 
The U.S. Government is authorized to reproduce and distribute reprints for government purposes not withstanding any copyright annotation hereon.

\section{Ethical Consideration}
This paper proposes a label selection method for weakly supervised text classification frameworks.
The aim of the paper is to detect the noise caused by the heuristic pseudo-labels and we don't intend to introduce any biased selection.
Based on our experiments, we manually inspected some filtered samples and we didn't find any underlying pattern.
Hence, we do not anticipate any major ethical concerns. 

\bibliography{ref}

\begin{thebibliography}{44}
\expandafter\ifx\csname natexlab\endcsname\relax\def\natexlab#1{#1}\fi

\bibitem[{Agichtein and Gravano(2000)}]{agichtein2000snowball}
Eugene Agichtein and Luis Gravano. 2000.
\newblock Snowball: Extracting relations from large plain-text collections.
\newblock In \emph{Proceedings of the fifth ACM conference on Digital
  libraries}, pages 85--94. ACM.

\bibitem[{Arachie and Huang(2021)}]{arachie2021constrained}
Chidubem Arachie and Bert Huang. 2021.
\newblock Constrained labeling for weakly supervised learning.
\newblock In \emph{Uncertainty in Artificial Intelligence}, pages 236--246.
  PMLR.

\bibitem[{Arpit et~al.(2017)Arpit, Jastrz{\k{e}}bski, Ballas, Krueger, Bengio,
  Kanwal, Maharaj, Fischer, Courville, Bengio et~al.}]{arpit2017closer}
Devansh Arpit, Stanis{\l}aw Jastrz{\k{e}}bski, Nicolas Ballas, David Krueger,
  Emmanuel Bengio, Maxinder~S Kanwal, Tegan Maharaj, Asja Fischer, Aaron
  Courville, Yoshua Bengio, et~al. 2017.
\newblock A closer look at memorization in deep networks.
\newblock In \emph{International Conference on Machine Learning}, pages
  233--242. PMLR.

\bibitem[{Corbi{\`e}re et~al.(2019)Corbi{\`e}re, Thome, Bar-Hen, Cord, and
  P{\'e}rez}]{Corbire2019AddressingFP}
Charles Corbi{\`e}re, Nicolas Thome, Avner Bar-Hen, Matthieu Cord, and Patrick
  P{\'e}rez. 2019.
\newblock Addressing failure prediction by learning model confidence.
\newblock In \emph{NeurIPS}.

\bibitem[{Devlin et~al.(2019)Devlin, Chang, Lee, and
  Toutanova}]{devlin2018bert}
Jacob Devlin, Ming-Wei Chang, Kenton Lee, and Kristina Toutanova. 2019.
\newblock \href {https://doi.org/10.18653/v1/N19-1423} {{BERT}: Pre-training of
  deep bidirectional transformers for language understanding}.
\newblock In \emph{Proceedings of the 2019 Conference of the North {A}merican
  Chapter of the Association for Computational Linguistics: Human Language
  Technologies, Volume 1 (Long and Short Papers)}, pages 4171--4186,
  Minneapolis, Minnesota. Association for Computational Linguistics.

\bibitem[{Devries and Taylor(2018)}]{Devries2018LearningCF}
Terrance Devries and Graham~W. Taylor. 2018.
\newblock Learning confidence for out-of-distribution detection in neural
  networks.
\newblock \emph{ArXiv}, abs/1802.04865.

\bibitem[{Dong et~al.(2021)Dong, Liu, and Shang}]{Dong2021DataQM}
Chengyu Dong, Liyuan Liu, and Jingbo Shang. 2021.
\newblock \href {http://arxiv.org/abs/2102.07437} {Data profiling for
  adversarial training: On the ruin of problematic data}.
\newblock \emph{CoRR}, abs/2102.07437.

\bibitem[{Fang et~al.(2020)Fang, Lu, Niu, and Sugiyama}]{fang2020rethinking}
Tongtong Fang, Nan Lu, Gang Niu, and Masashi Sugiyama. 2020.
\newblock \href
  {https://proceedings.neurips.cc/paper/2020/file/8b9e7ab295e87570551db122a04c6f7c-Paper.pdf}
  {Rethinking importance weighting for deep learning under distribution shift}.
\newblock In \emph{Advances in Neural Information Processing Systems},
  volume~33, pages 11996--12007. Curran Associates, Inc.

\bibitem[{Geifman et~al.(2019)Geifman, Uziel, and El-Yaniv}]{geifman2018bias}
Yonatan Geifman, Guy Uziel, and Ran El-Yaniv. 2019.
\newblock \href {https://openreview.net/forum?id=SJfb5jCqKm} {Bias-reduced
  uncertainty estimation for deep neural classifiers}.
\newblock In \emph{International Conference on Learning Representations}.

\bibitem[{Guo et~al.(2017)Guo, Pleiss, Sun, and
  Weinberger}]{guo2017calibration}
Chuan Guo, Geoff Pleiss, Yu~Sun, and Kilian~Q Weinberger. 2017.
\newblock On calibration of modern neural networks.
\newblock In \emph{International Conference on Machine Learning}, pages
  1321--1330. PMLR.

\bibitem[{Hacohen et~al.(2020)Hacohen, Choshen, and
  Weinshall}]{Hacohen2019LetsAT}
Guy Hacohen, Leshem Choshen, and Daphna Weinshall. 2020.
\newblock \href {http://proceedings.mlr.press/v119/hacohen20a.html} {Let's
  agree to agree: Neural networks share classification order on real datasets}.
\newblock In \emph{ICML}, pages 3950--3960.

\bibitem[{Han et~al.(2018)Han, Yao, Yu, Niu, Xu, Hu, Tsang, and
  Sugiyama}]{han2018co}
Bo~Han, Quanming Yao, Xingrui Yu, Gang Niu, Miao Xu, Weihua Hu, Ivor Tsang, and
  Masashi Sugiyama. 2018.
\newblock Co-teaching: Robust training of deep neural networks with extremely
  noisy labels.
\newblock \emph{Advances in neural information processing systems}, 31.

\bibitem[{Hecker et~al.(2018)Hecker, Dai, and Gool}]{Hecker2018FailurePF}
Simon Hecker, Dengxin Dai, and Luc~Van Gool. 2018.
\newblock Failure prediction for autonomous driving.
\newblock \emph{2018 IEEE Intelligent Vehicles Symposium (IV)}, pages
  1792--1799.

\bibitem[{Hendrycks and Gimpel(2017)}]{Hendrycks2017ABF}
Dan Hendrycks and Kevin Gimpel. 2017.
\newblock A baseline for detecting misclassified and out-of-distribution
  examples in neural networks.
\newblock \emph{Proceedings of International Conference on Learning
  Representations}.

\bibitem[{Huang et~al.(2019)Huang, Qu, Jia, and Zhao}]{huang2019o2u}
Jinchi Huang, Lie Qu, Rongfei Jia, and Binqiang Zhao. 2019.
\newblock O2u-net: A simple noisy label detection approach for deep neural
  networks.
\newblock In \emph{Proceedings of the IEEE/CVF International Conference on
  Computer Vision}, pages 3326--3334.

\bibitem[{Jiang et~al.(2018{\natexlab{a}})Jiang, Kim, and
  Gupta}]{Jiang2018ToTO}
Heinrich Jiang, Been Kim, and Maya~R. Gupta. 2018{\natexlab{a}}.
\newblock To trust or not to trust a classifier.
\newblock In \emph{NeurIPS}.

\bibitem[{Jiang et~al.(2018{\natexlab{b}})Jiang, Zhou, Leung, Li, and
  Fei-Fei}]{jiang2018mentornet}
Lu~Jiang, Zhengyuan Zhou, Thomas Leung, Li-Jia Li, and Li~Fei-Fei.
  2018{\natexlab{b}}.
\newblock Mentornet: Learning data-driven curriculum for very deep neural
  networks on corrupted labels.
\newblock In \emph{International Conference on Machine Learning}, pages
  2304--2313. PMLR.

\bibitem[{Karamanolakis et~al.(2021{\natexlab{a}})Karamanolakis, Mukherjee,
  Zheng, and Awadallah}]{karamanolakis-etal-2021-self}
Giannis Karamanolakis, Subhabrata Mukherjee, Guoqing Zheng, and Ahmed~Hassan
  Awadallah. 2021{\natexlab{a}}.
\newblock \href {https://doi.org/10.18653/v1/2021.naacl-main.66} {Self-training
  with weak supervision}.
\newblock In \emph{Proceedings of the 2021 Conference of the North American
  Chapter of the Association for Computational Linguistics: Human Language
  Technologies}, pages 845--863, Online. Association for Computational
  Linguistics.

\bibitem[{Karamanolakis et~al.(2021{\natexlab{b}})Karamanolakis, Mukherjee,
  Zheng, and Awadallah}]{karamanolakis2021self-training}
Giannis Karamanolakis, Subhabrata~(Subho) Mukherjee, Guoqing Zheng, and
  Ahmed~H. Awadallah. 2021{\natexlab{b}}.
\newblock \href
  {https://www.microsoft.com/en-us/research/publication/self-training-weak-supervision-astra/}
  {Self-training with weak supervision}.
\newblock In \emph{NAACL 2021}. NAACL 2021.

\bibitem[{Lee et~al.(2018)Lee, Lee, Lee, and Shin}]{Lee2018ASU}
Kimin Lee, Kibok Lee, Honglak Lee, and Jinwoo Shin. 2018.
\newblock A simple unified framework for detecting out-of-distribution samples
  and adversarial attacks.
\newblock In \emph{NeurIPS}.

\bibitem[{Liang et~al.(2018)Liang, Li, and Srikant}]{Liang2018EnhancingTR}
Shiyu Liang, Yixuan Li, and R.~Srikant. 2018.
\newblock \href {https://openreview.net/forum?id=H1VGkIxRZ} {Enhancing the
  reliability of out-of-distribution image detection in neural networks}.
\newblock In \emph{International Conference on Learning Representations}.

\bibitem[{Liu et~al.(2019)Liu, Ott, Goyal, Du, Joshi, Chen, Levy, Lewis,
  Zettlemoyer, and Stoyanov}]{liu2019roberta}
Yinhan Liu, Myle Ott, Naman Goyal, Jingfei Du, Mandar Joshi, Danqi Chen, Omer
  Levy, Mike Lewis, Luke Zettlemoyer, and Veselin Stoyanov. 2019.
\newblock Roberta: A robustly optimized bert pretraining approach.
\newblock \emph{arXiv preprint arXiv:1907.11692}.

\bibitem[{Malach and Shalev-Shwartz(2017)}]{malach2017decoupling}
Eran Malach and Shai Shalev-Shwartz. 2017.
\newblock \href
  {https://proceedings.neurips.cc/paper/2017/file/58d4d1e7b1e97b258c9ed0b37e02d087-Paper.pdf}
  {Decoupling "when to update" from "how to update"}.
\newblock In \emph{Advances in Neural Information Processing Systems},
  volume~30. Curran Associates, Inc.

\bibitem[{Mekala et~al.(2021)Mekala, Gangal, and Shang}]{mekala2021coarse2fine}
Dheeraj Mekala, Varun Gangal, and Jingbo Shang. 2021.
\newblock \href {https://doi.org/10.18653/v1/2021.emnlp-main.46}
  {{C}oarse2{F}ine: Fine-grained text classification on coarsely-grained
  annotated data}.
\newblock In \emph{Proceedings of the 2021 Conference on Empirical Methods in
  Natural Language Processing}, pages 583--594, Online and Punta Cana,
  Dominican Republic. Association for Computational Linguistics.

\bibitem[{Mekala and Shang(2020)}]{mekala2020contextualized}
Dheeraj Mekala and Jingbo Shang. 2020.
\newblock Contextualized weak supervision for text classification.
\newblock In \emph{Proceedings of the 58th Annual Meeting of the Association
  for Computational Linguistics}, pages 323--333.

\bibitem[{Mekala et~al.(2020)Mekala, Zhang, and Shang}]{mekala2020meta}
Dheeraj Mekala, Xinyang Zhang, and Jingbo Shang. 2020.
\newblock Meta: Metadata-empowered weak supervision for text classification.
\newblock In \emph{Proceedings of the 2020 Conference on Empirical Methods in
  Natural Language Processing (EMNLP)}, pages 8351--8361.

\bibitem[{Meng et~al.(2018)Meng, Shen, Zhang, and Han}]{meng2018weakly}
Yu~Meng, Jiaming Shen, Chao Zhang, and Jiawei Han. 2018.
\newblock Weakly-supervised neural text classification.
\newblock In \emph{Proceedings of the 27th ACM International Conference on
  Information and Knowledge Management}, pages 983--992. ACM.

\bibitem[{Meng et~al.(2019)Meng, Shen, Zhang, and Han}]{meng2019weakly}
Yu~Meng, Jiaming Shen, Chao Zhang, and Jiawei Han. 2019.
\newblock Weakly-supervised hierarchical text classification.
\newblock In \emph{Proceedings of the AAAI Conference on Artificial
  Intelligence}, volume~33, pages 6826--6833.

\bibitem[{Meng et~al.(2020)Meng, Zhang, Huang, Xiong, Ji, Zhang, and
  Han}]{meng2020text}
Yu~Meng, Yunyi Zhang, Jiaxin Huang, Chenyan Xiong, Heng Ji, Chao Zhang, and
  Jiawei Han. 2020.
\newblock Text classification using label names only: A language model
  self-training approach.
\newblock In \emph{Proceedings of the 2020 Conference on Empirical Methods in
  Natural Language Processing}.

\bibitem[{Mukherjee and Awadallah(2020)}]{mukherjee2020uncertainty}
Subhabrata Mukherjee and Ahmed Awadallah. 2020.
\newblock Uncertainty-aware self-training for few-shot text classification.
\newblock \emph{Advances in Neural Information Processing Systems},
  33:21199--21212.

\bibitem[{Radford et~al.(2019)Radford, Wu, Child, Luan, Amodei, Sutskever
  et~al.}]{radford2019language}
Alec Radford, Jeffrey Wu, Rewon Child, David Luan, Dario Amodei, Ilya
  Sutskever, et~al. 2019.
\newblock Language models are unsupervised multitask learners.
\newblock \emph{OpenAI blog}, 1(8):9.

\bibitem[{Ren et~al.(2018)Ren, Zeng, Yang, and Urtasun}]{ren2018learning}
Mengye Ren, Wenyuan Zeng, Bin Yang, and Raquel Urtasun. 2018.
\newblock Learning to reweight examples for robust deep learning.
\newblock In \emph{International Conference on Machine Learning}, pages
  4334--4343. PMLR.

\bibitem[{Riloff et~al.(2003)Riloff, Wiebe, and Wilson}]{riloff2003learning}
Ellen Riloff, Janyce Wiebe, and Theresa Wilson. 2003.
\newblock Learning subjective nouns using extraction pattern bootstrapping.
\newblock In \emph{Proceedings of the seventh conference on Natural language
  learning at HLT-NAACL 2003-Volume 4}, pages 25--32. Association for
  Computational Linguistics.

\bibitem[{Rizve et~al.(2021)Rizve, Duarte, Rawat, and Shah}]{rizve2021defense}
Mamshad~Nayeem Rizve, Kevin Duarte, Yogesh~S Rawat, and Mubarak Shah. 2021.
\newblock \href {https://openreview.net/forum?id=-ODN6SbiUU} {In defense of
  pseudo-labeling: An uncertainty-aware pseudo-label selection framework for
  semi-supervised learning}.
\newblock In \emph{International Conference on Learning Representations}.

\bibitem[{Swayamdipta et~al.(2020)Swayamdipta, Schwartz, Lourie, Wang,
  Hajishirzi, Smith, and Choi}]{swayamdipta-etal-2020-dataset}
Swabha Swayamdipta, Roy Schwartz, Nicholas Lourie, Yizhong Wang, Hannaneh
  Hajishirzi, Noah~A. Smith, and Yejin Choi. 2020.
\newblock \href {https://doi.org/10.18653/v1/2020.emnlp-main.746} {Dataset
  cartography: Mapping and diagnosing datasets with training dynamics}.
\newblock In \emph{Proceedings of the 2020 Conference on Empirical Methods in
  Natural Language Processing (EMNLP)}, pages 9275--9293, Online. Association
  for Computational Linguistics.

\bibitem[{Tao et~al.(2015)Tao, Zhang, Chen, Jiang, Hanratty, Kaplan, and
  Han}]{tao2015doc2cube}
Fangbo Tao, Chao Zhang, Xiusi Chen, Meng Jiang, Tim Hanratty, Lance Kaplan, and
  Jiawei Han. 2015.
\newblock Doc2cube: Automated document allocation to text cube via
  dimension-aware joint embedding.
\newblock \emph{Dimension}, 2016:2017.

\bibitem[{Toneva et~al.(2019)Toneva, Sordoni, des Combes, Trischler, Bengio,
  and Gordon}]{Toneva2019AnES}
Mariya Toneva, Alessandro Sordoni, Remi~Tachet des Combes, Adam Trischler,
  Yoshua Bengio, and Geoffrey~J. Gordon. 2019.
\newblock \href {https://openreview.net/forum?id=BJlxm30cKm} {An empirical
  study of example forgetting during deep neural network learning}.
\newblock In \emph{International Conference on Learning Representations}.

\bibitem[{Wan and McAuley(2018)}]{DBLP:conf/recsys/WanM18}
Mengting Wan and Julian~J. McAuley. 2018.
\newblock \href {https://doi.org/10.1145/3240323.3240369} {Item recommendation
  on monotonic behavior chains}.
\newblock In \emph{Proceedings of the 12th {ACM} Conference on Recommender
  Systems, RecSys 2018, Vancouver, BC, Canada, October 2-7, 2018}, pages
  86--94. {ACM}.

\bibitem[{Wan et~al.(2019)Wan, Misra, Nakashole, and
  McAuley}]{DBLP:conf/acl/WanMNM19}
Mengting Wan, Rishabh Misra, Ndapa Nakashole, and Julian~J. McAuley. 2019.
\newblock \href {https://doi.org/10.18653/v1/p19-1248} {Fine-grained spoiler
  detection from large-scale review corpora}.
\newblock In \emph{Proceedings of the 57th Conference of the Association for
  Computational Linguistics, {ACL} 2019, Florence, Italy, July 28- August 2,
  2019, Volume 1: Long Papers}, pages 2605--2610. Association for Computational
  Linguistics.

\bibitem[{Wang et~al.(2021)Wang, Mekala, and Shang}]{wang2020x}
Zihan Wang, Dheeraj Mekala, and Jingbo Shang. 2021.
\newblock \href {https://doi.org/10.18653/v1/2021.naacl-main.242} {{X}-class:
  Text classification with extremely weak supervision}.
\newblock In \emph{Proceedings of the 2021 Conference of the North American
  Chapter of the Association for Computational Linguistics: Human Language
  Technologies}, pages 3043--3053, Online. Association for Computational
  Linguistics.

\bibitem[{Yang et~al.(2019)Yang, Dai, Yang, Carbonell, Salakhutdinov, and
  Le}]{yang2019xlnet}
Zhilin Yang, Zihang Dai, Yiming Yang, Jaime Carbonell, Russ~R Salakhutdinov,
  and Quoc~V Le. 2019.
\newblock Xlnet: Generalized autoregressive pretraining for language
  understanding.
\newblock \emph{Advances in neural information processing systems}, 32.

\bibitem[{Yu et~al.(2019)Yu, Han, Yao, Niu, Tsang, and Sugiyama}]{yu2019does}
Xingrui Yu, Bo~Han, Jiangchao Yao, Gang Niu, Ivor Tsang, and Masashi Sugiyama.
  2019.
\newblock How does disagreement help generalization against label corruption?
\newblock In \emph{International Conference on Machine Learning}, pages
  7164--7173. PMLR.

\bibitem[{Zhang et~al.(2021)Zhang, Bengio, Hardt, Recht, and
  Vinyals}]{zhang2021understanding}
Chiyuan Zhang, Samy Bengio, Moritz Hardt, Benjamin Recht, and Oriol Vinyals.
  2021.
\newblock Understanding deep learning (still) requires rethinking
  generalization.
\newblock \emph{Communications of the ACM}, 64(3):107--115.

\bibitem[{Zhang et~al.(2015)Zhang, Zhao, and LeCun}]{zhang2015character}
Xiang Zhang, Junbo Zhao, and Yann LeCun. 2015.
\newblock Character-level convolutional networks for text classification.
\newblock \emph{Advances in neural information processing systems},
  28:649--657.

\end{thebibliography}
\bibliographystyle{acl_natbib}

\clearpage
\newpage
\appendix
\section{Appendix}
\label{sec:appendix}

\begin{figure}[t]
    \subfigure[]{
        \includegraphics[width=0.45\linewidth]{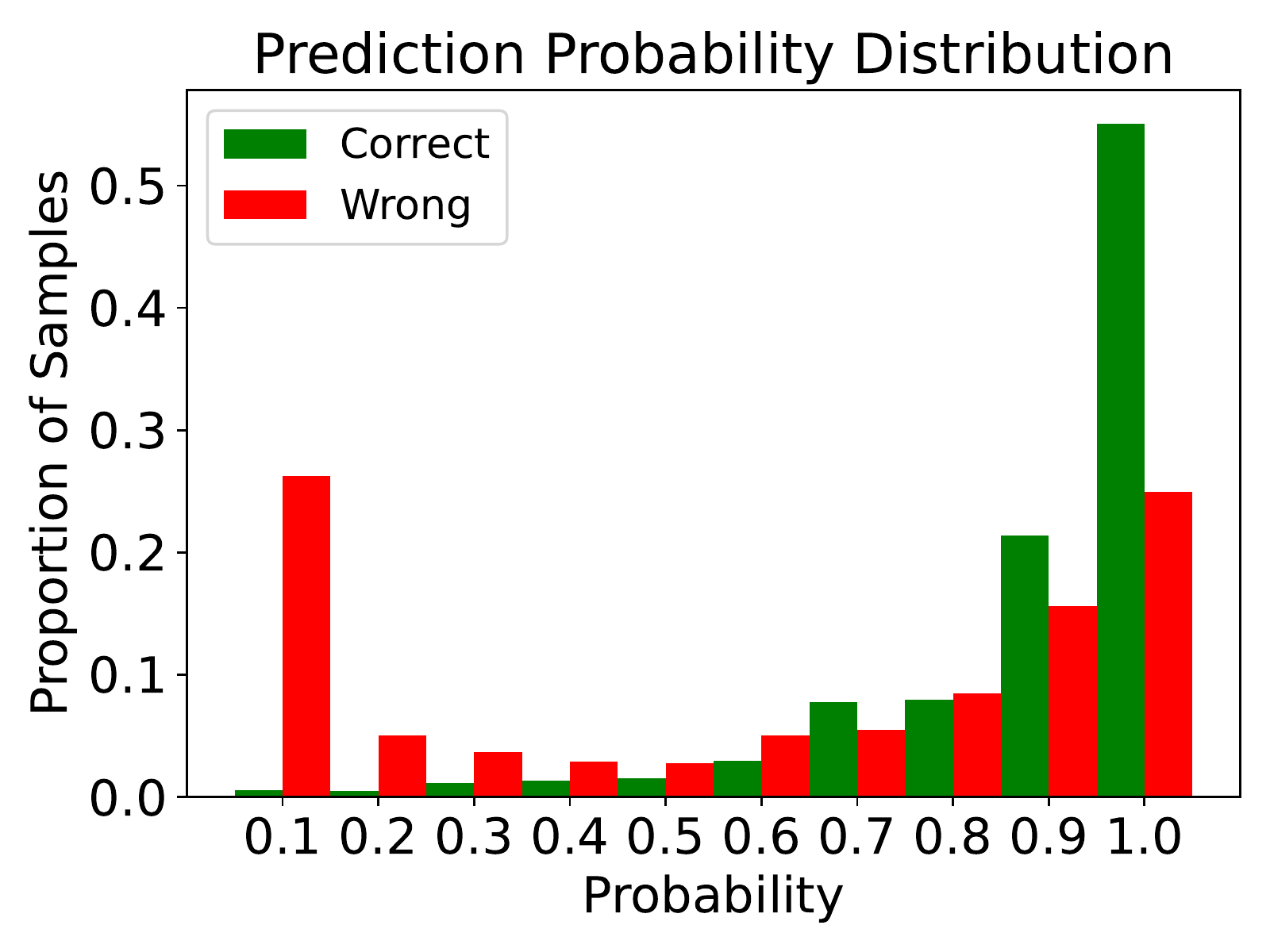}
        \label{figure:prob_filter_nyt-fine}
    }
    \subfigure[]{
        \includegraphics[width=0.45\linewidth]{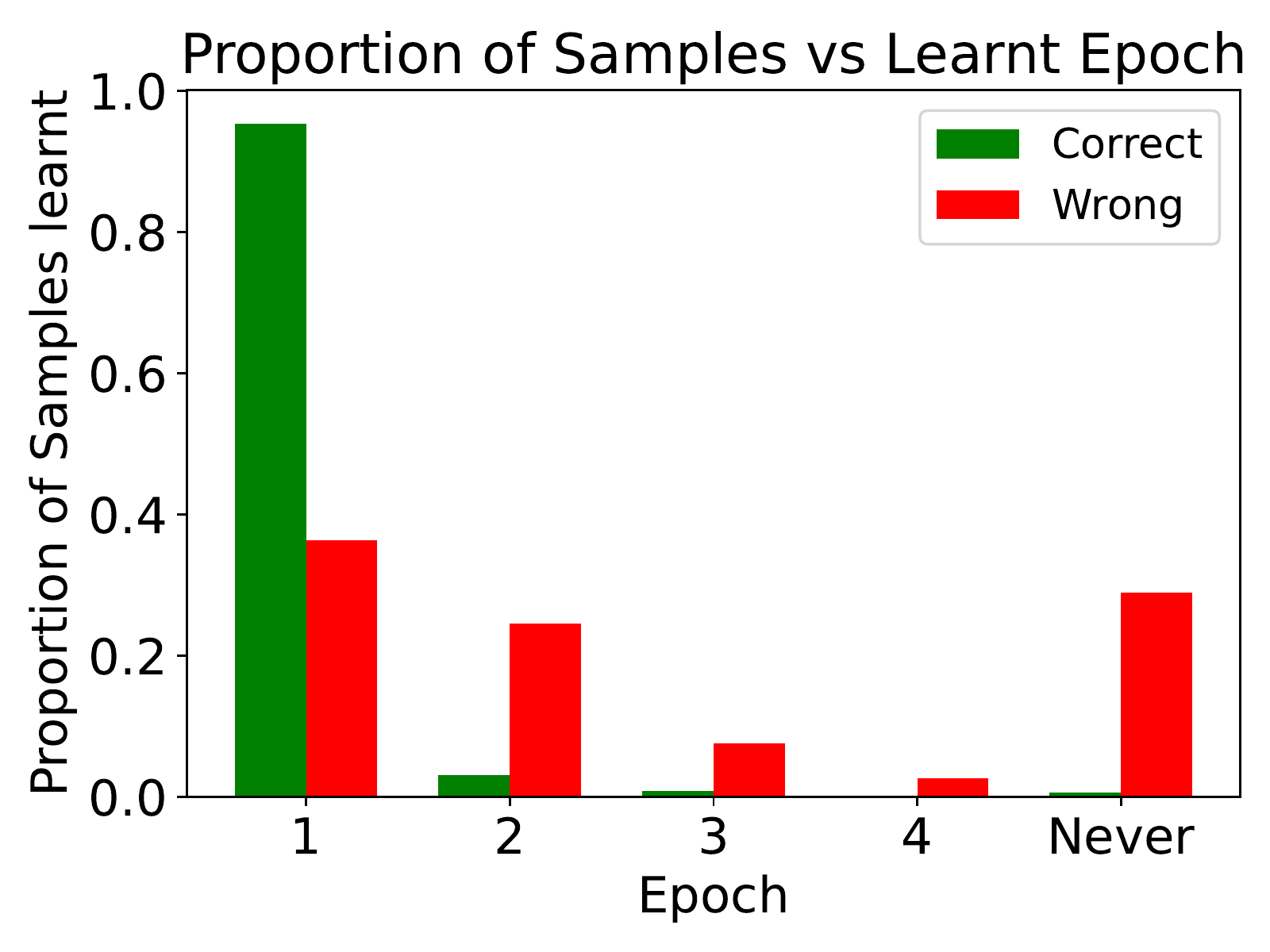}
        \label{figure:ep_filter_nyt-fine}
    }
    \subfigure[]{
        \includegraphics[width=0.45\linewidth]{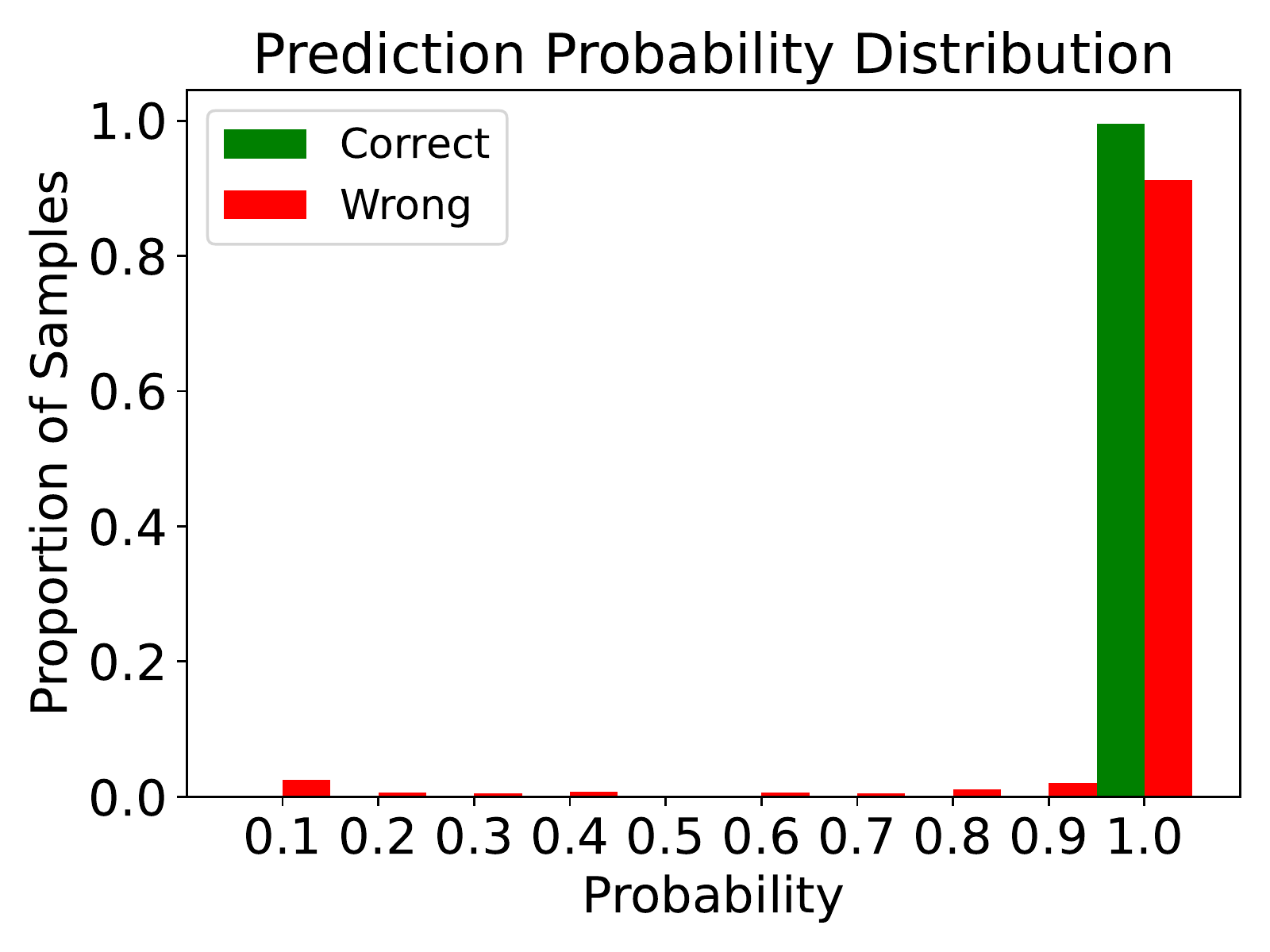}
        \label{figure:prob_filter_20news-coarse}
    }
    \subfigure[]{
        \includegraphics[width=0.45\linewidth]{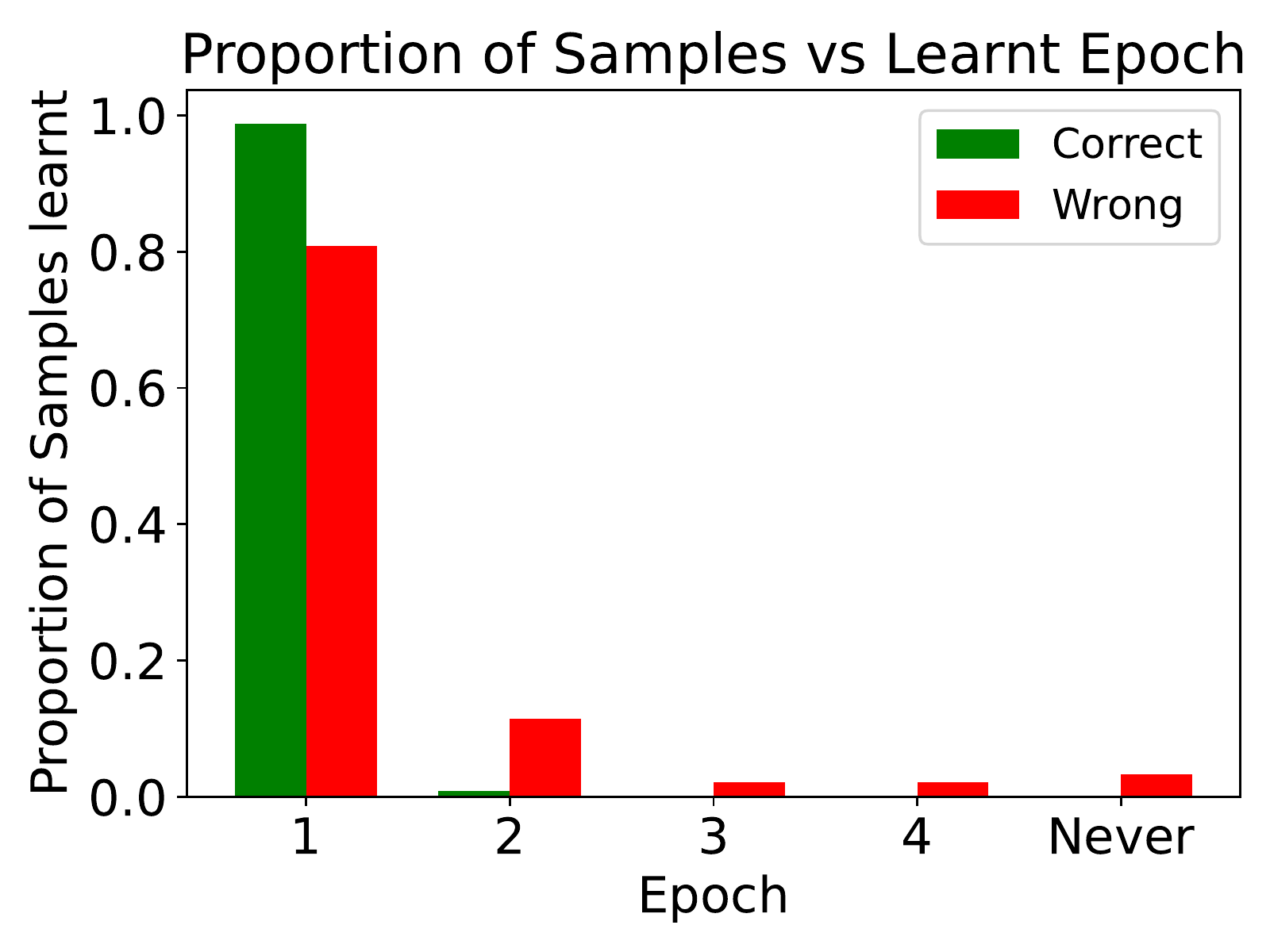}
        \label{figure:ep_filter_20news-coarse}
    }
    \subfigure[]{
        \includegraphics[width=0.45\linewidth]{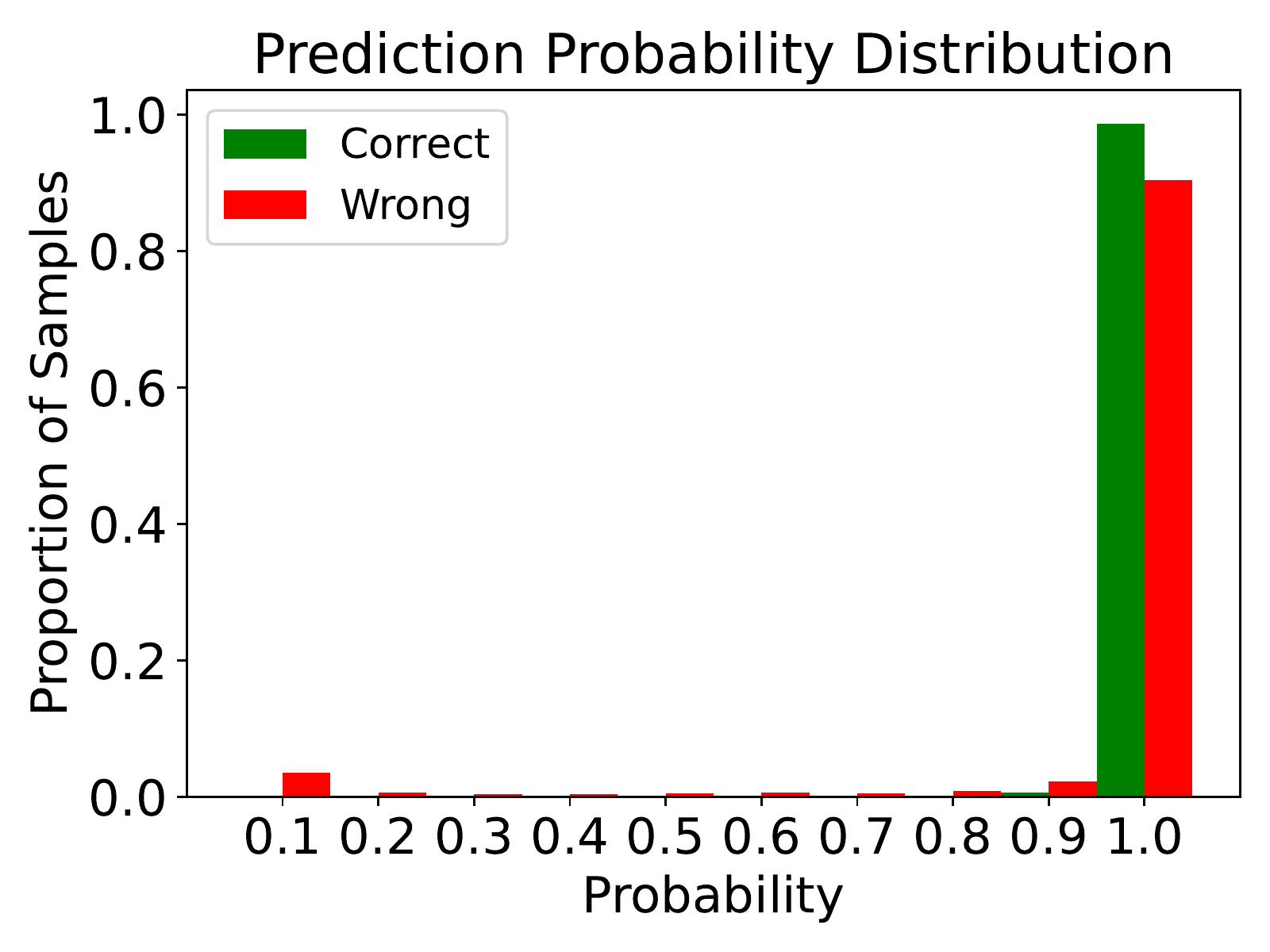}
        \label{figure:prob_filter_20news-fine}
    }
    \subfigure[]{
        \includegraphics[width=0.45\linewidth]{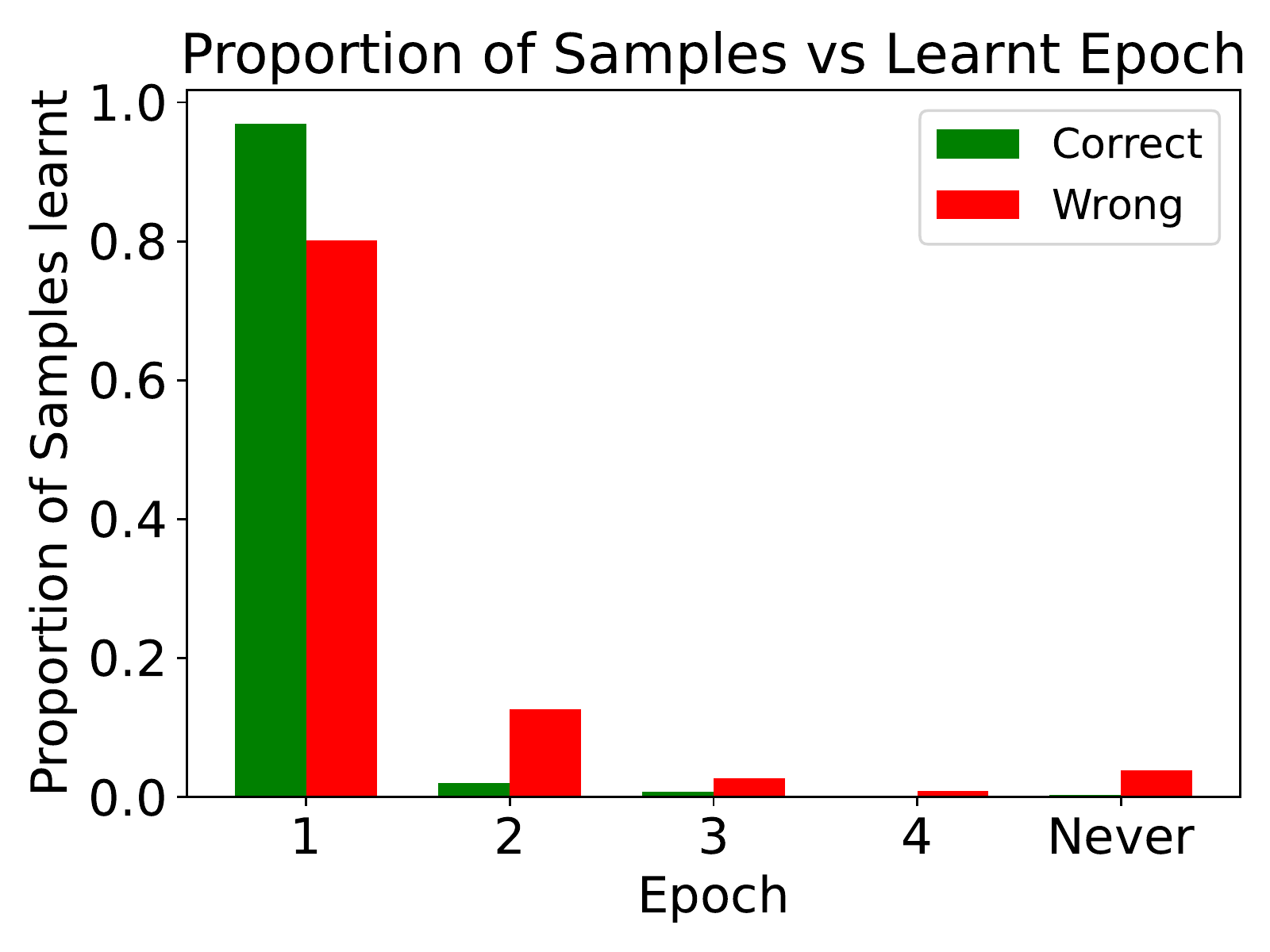}
        \label{figure:ep_filter_20news-fine}
    }
    \subfigure[]{
        \includegraphics[width=0.45\linewidth]{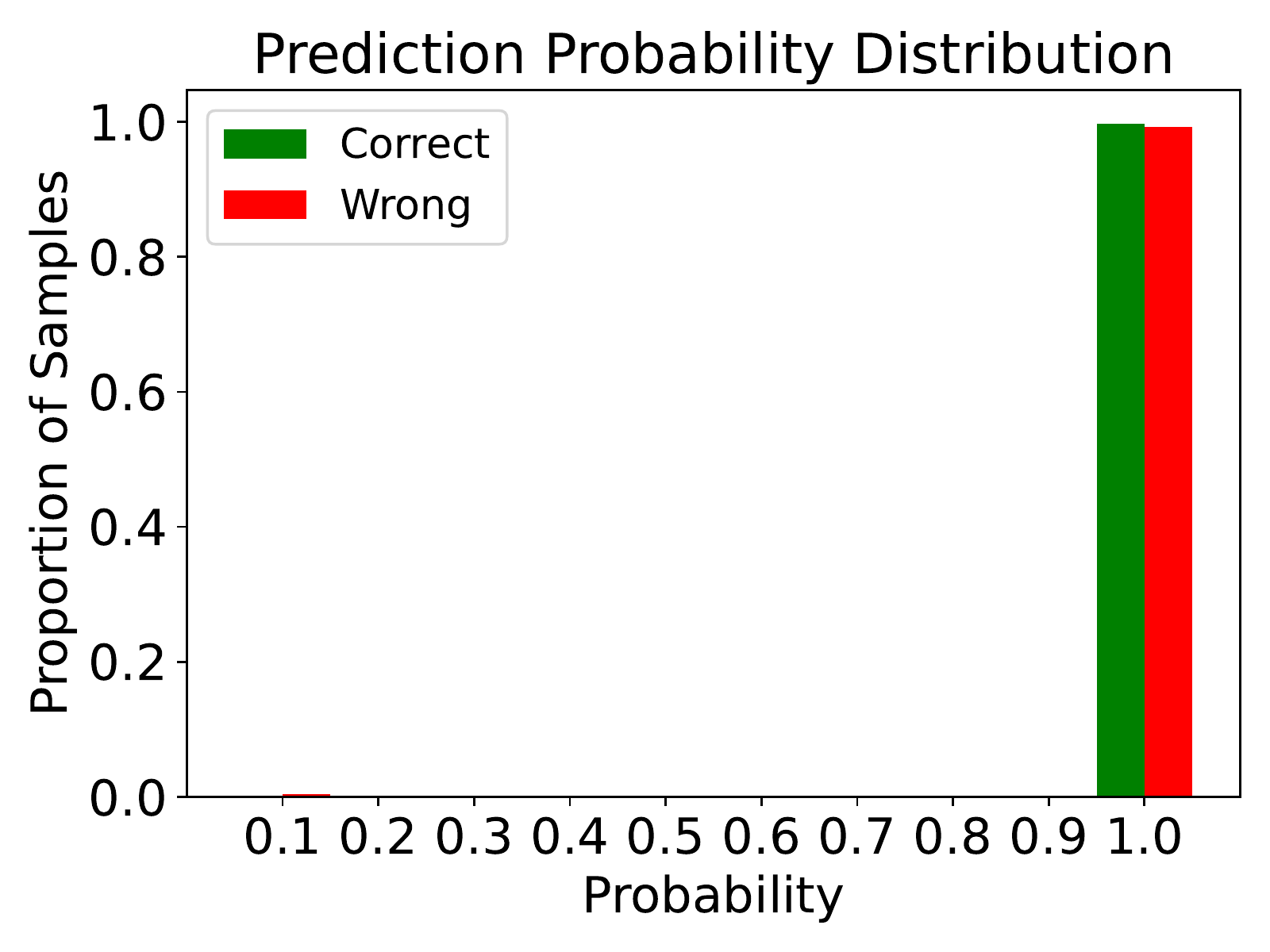}
        \label{figure:prob_filter_books}
    }
    \subfigure[]{
        \includegraphics[width=0.45\linewidth]{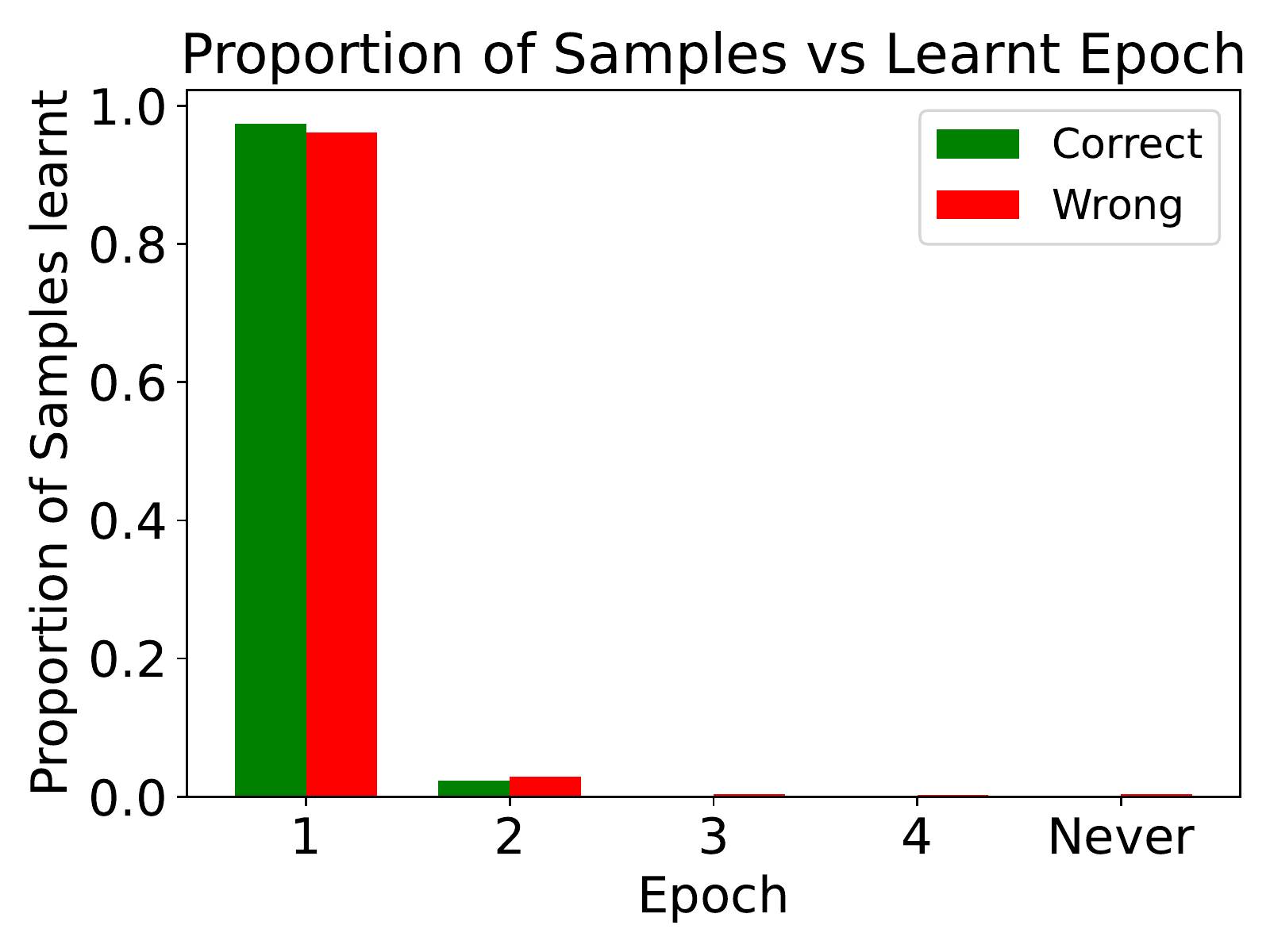}
        \label{figure:ep_filter_books}
    }
    \subfigure[]{
        \includegraphics[width=0.45\linewidth]{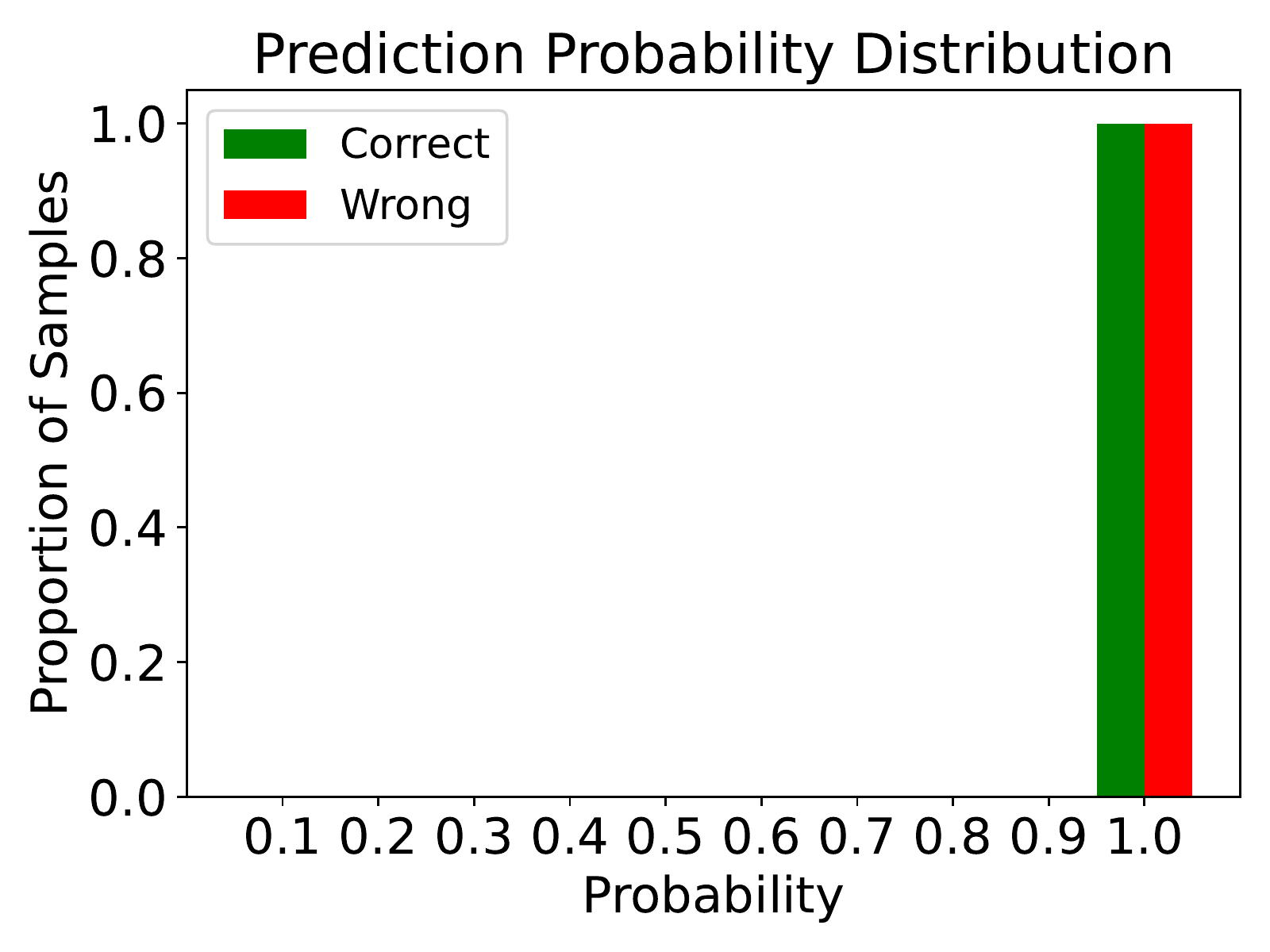}
        \label{figure:prob_filter_agnews}
    }
    \subfigure[]{
        \includegraphics[width=0.45\linewidth]{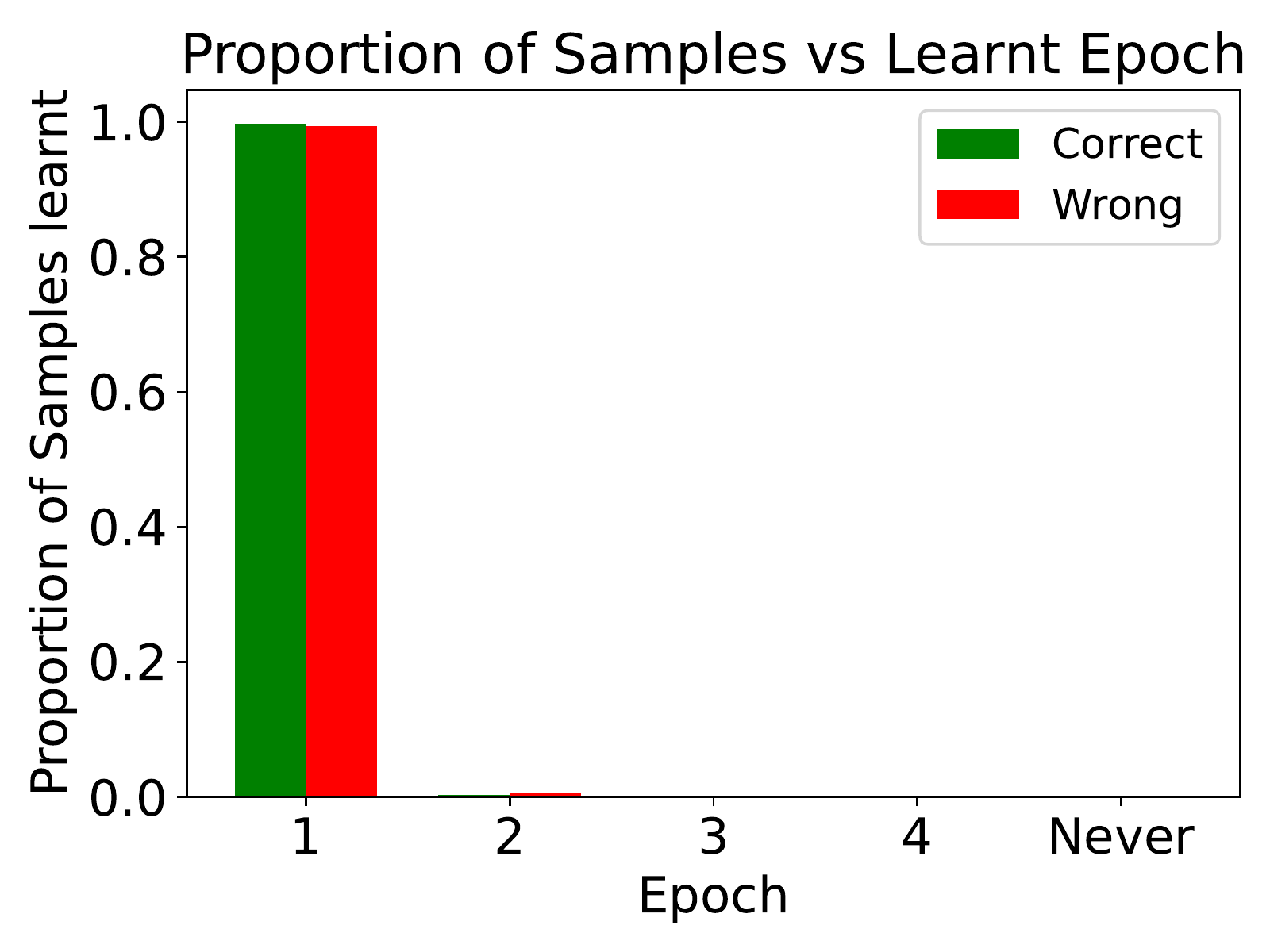}
        \label{figure:ep_filter_agnews}
    }
    \vspace{-3mm}
    \caption{
        Distributions of correctly and wrongly labeled pseudo-labels using different selection strategies on all datasets for its initial pseudo-labels.
        The base classifier is BERT. Each row represents a dataset. Figure
        (a), (b) represents NYT-Fine, (c), (d) represents 20News-Coarse, (e), (f) represents 20News-Fine, (g), (h) represents Books, and (i), (j) represents AGNews datasets respectively.
        Left column is based on the softmax probability of samples’ pseudo-labels and right column is based on the earliest epochs at which samples are learnt.
    }
    \label{figure:prelim_exp}
    \vspace{-5mm}
\end{figure}

\subsection{Datasets}
\label{sec:datasets}
The details of datasets are provided below:
\begin{itemize}[leftmargin=*,nosep]
    \item \textbf{The New York Times (NYT):} The NYT dataset is a collection of news articles published by The New York Times. They are classified into 5 coarse-grained genres (e.g., science, sports) and 25 fine-grained categories (e.g., music, football, dance, basketball).
    \item \textbf{The 20 Newsgroups (20News):} The 20News dataset\footnote{\url{http://qwone.com/~jason/20Newsgroups/}} is a collection of newsgroup documents partitioned widely into 6 groups (e.g., recreation, computers) and 20 fine-grained classes (e.g., graphics, windows, baseball, hockey). Following ~\cite{wang2020x}, coarse- and fine-grained miscellaneous labels are ignored.
    \item \textbf{AGNews}~\cite{zhang2015character} is a huge collection of news articles categorized into four coarse-grained topics such as business, politics, sports, and technology.
    \item \textbf{Books}~\cite{DBLP:conf/recsys/WanM18, DBLP:conf/acl/WanMNM19} is a dataset containing description of books, user-book interactions, and users’ book reviews collected from a popular online book review website Goodreads\footnote{\url{https://www.goodreads.com/}}. Following ~\cite{mekala2020meta}, we select books belonging to eight popular genres. Using the title and description as text, we aim to predict the genre of a book.
\end{itemize}

\subsection{Compared Weakly Supervised Text Classification Methods}
\label{sec:methods}

We compared with following state-of-the-art weakly supervised text classification methods described below\footnote{We also considered experimenting on ASTRA, however the instructions to run on custom datasets were not made public yet.}:
\begin{itemize}[leftmargin=*,nosep]
    \item \textbf{ConWea}~\cite{mekala2020contextualized} is a seed-word driven iterative framework that uses pre-trained language models to contextualize the weak supervision.
    \item \textbf{X-Class}~\cite{wang2020x} takes only label surface names as supervision and learns class-oriented document representations. These document representations are aligned to classes, computing pseudo labels for training a classifier.
    \item \textbf{WeSTClass}~\cite{meng2018weakly} generates pseudo documents using seed information and refines the model through a self-training module that bootstraps on unlabeled documents.
    \item \textbf{LOTClass}~\cite{meng2020text} queries replacements of class names using BERT~\cite{devlin2018bert} and constructs a category vocabulary for each class. This is used to pseudo-label the documents via string matching. A classifier is trained on this pseudo-labeled data with further self-training.
\end{itemize}

We use the public implementations of these methods and modify them to plug-in our filter. 
Specifically, in WeSTClass and LOTClass, we add our filter after generating the pseudo documents; in ConWea, we add our filter before training the text classifier; and for X-Class, we plug-in our filter after learning the document-class alignment.

\subsection{Experimental Settings}
\smallsection{Train-Test sets}
We remove the labels in the whole dataset and our task is to assign labels to these unlabeled samples. We measure our performance on the whole dataset by comparing it with their respective gold labels.

\smallsection{Computation Infrastructure}
We performed our experiments on NVIDIA RTX A6000 GPU. The batch size for training BERT is 32, RoBERTa is 32, GPT2 is 4, XLNet is 1. The running time for BERT and RoBERTa took 3 hrs, GPT2 took 6 hours, and XLNet took 12 hrs.


\subsection{Additional Experiments}
We also compare with \textbf{RoBERTa} (\verb+roberta-base+)~\cite{liu2019roberta} as text classifier. We fine-tune it for 3 epochs. The results are shown in Table~\ref{results-add}.
\begin{table*}[t]
\centering
\caption{Evaluation results on six datasets using RoBERTa classifier and pseudo-label selection methods. 
Initial pseudo-labels are generated using String-Match. 
Micro- and Macro-F1 scores and their respective standard deviations are presented in percentages. 
For a fair comparison, we consider the same number of samples for all baselines as \our in each iteration.
Abnormally high standard deviations are highlighted in \textcolor{blue}{\textit{blue}} and low performances are highlighted in \textcolor{red}{\textit{red}}. Baselines performing better than our method are made bold.
}
\small
\resizebox{\linewidth}{!}{
\setlength{\tabcolsep}{1mm}
\begin{tabular}{c c cc cc cc cc cc cc}
\toprule
 & & \multicolumn{8}{c}{Coarse-grained Datasets} & \multicolumn{4}{c}{Fine-grained Datasets} \\
 \cmidrule(lr){3-10} \cmidrule(lr){11-14}
 & & \multicolumn{2}{c}{NYT-Coarse} & \multicolumn{2}{c}{20News-Coarse} & \multicolumn{2}{c}{AGNews} & \multicolumn{2}{c}{Books} & \multicolumn{2}{c}{NYT-Fine} & \multicolumn{2}{c}{20News-Fine}\\
 \cmidrule(lr){3-4} \cmidrule(lr){5-6} \cmidrule(lr){7-8} \cmidrule(lr){9-10} \cmidrule(lr){11-12} \cmidrule(lr){13-14}
Classifier & Method & Mi-F1 & Ma-F1 & Mi-F1 & Ma-F1 & Mi-F1 & Ma-F1 & Mi-F1 & Ma-F1 & Mi-F1 & Ma-F1 & Mi-F1 & Ma-F1 \\
\midrule
\multirow{7}{*}{RoBERTa} & Standard & 90.2(0.41) & 82.1(0.24) & 76.5(0.41) & 75.7(0.58) & 74.4(0.44) & 74.2(0.71) & 57.6(0.29) & 58.6(0.53) & 79.4(0.65) & 76.6(0.54) & 67.4(0.67) & 67.3(0.87)\\
                        & \our & \textbf{92.4(2.99)} & \textbf{85.6(3.00)} & \textbf{77.5(2.00)} & \textbf{75.8(2.00)} & \textbf{75.6(0.22)} & \textbf{75.5(0.27)} & \textbf{59.7(0.41)} & \textbf{60.5(0.45)} & \textbf{81.8(0.90)} & \textbf{80.7(0.50)} & \textbf{70.7(0.68)} & \textbf{70.8(0.34)}\\
\cmidrule{2-14}
                        & O2U-Net & 93.1(0.14) & 86.3(0.26) & 76.5(0.19) & 73.4(1.47) & \textbf{77.6(0.36)} & \textbf{77.1(0.54)}  & 58.5(0.64) & 59.9(0.32) & 79.2(0.28) & 77.5(1.17) & 68.4(0.47) & 68.3(0.15)\\
                      & Random & 92.3(0.21) & 84.4(0.82) & 76.5(1.00) & 74.5(1.00) & 74.6(0.32) & 74.2(0.27) & 56.4(0.57) & 58.7(0.32) & 76.6(1.25) & 74.8(0.34) & 68.4(0.23) & 68.5(0.23)\\
                      & Probability & \textbf{93.4(0.48)} & \textbf{87.5(1.00)} & 76.7(0.50) & 75.4(1.00) & 76.2(0.89) & 76.3(1.12) & 56.2(1.28) & 57.4(1.85) & \textcolor{red}{26.6(23.00)} & \textcolor{red}{14.4(11.50)} & \textcolor{blue}{46.2(23.00)} & \textcolor{blue}{45.3(23.50)}\\
                      & Stability & 90.5(1.09) & 83.3(0.50) & \textbf{78.5(1.00)} & \textbf{76.0(1.50)} & 76.5(0.48) & 76.5(0.64) & 58.5(1.18) & 59.5(1.06) & \textcolor{red}{21.5(12.50)} & \textcolor{red}{9.2(5.00)} & 70.3(1.00) & 70.6(1.00)\\
\cmidrule{2-14}
                      & OptimalFilter & 98.2(0.17) & 96.1(0.16) & 94.3(0.74) & 94.5(0.35) & 89.7(0.17) & 89.3(0.28) & 76.5(0.29) & 77.7(0.22) & 97.4(0.34) & 92.8(0.26) & 85.3(0.32) & 85.5(0.65) \\
\bottomrule
\end{tabular}
}
\label{results-add}
\end{table*}

\subsection{Statistical Significance Tests}
\label{sec:statsig}
We perform a paired t-test between LOPS and each of the other baseline filtering techniques for all classifiers and on all datasets. The results are showed in Table~\ref{results-statsig}. From these p-values, we can conclude that the performance improvement over baselines is significant.

\begin{table*}[t]
\centering
\caption{Statistical significance results.}
\scalebox{0.7}{
\begin{tabular}{c|c|c|c|c|c|c|c|c|c|c|c|c|c}
\hline
Classifier & Method & NYT-Coarse & NYT-Fine & 20News-Coarse & 20News-Fine & AGNews & Books\\\hline
\multirow{6}{*}{BERT} & Standard & $1.93$ $\times$ $10^{-112}$ & $1.92$ $\times$ $10^{-105}$  & $7.08$ $\times$ $10^{-80}$  & $9.37$ $\times$ $10^{-79}$ & $1.05$ $\times$ $10^{-74}$  & $7.15$ $\times$ $10^{-96}$ \\\cline{2-8}
                      & Random & $1.58$ $\times$ $10^{-115}$ & $2.01$ $\times$ $10^{-105}$ & $5.98$ $\times$ $10^{-94}$ & $7.32$ $\times$ $10^{-39}$ & $4.26$ $\times$ $10^{-81}$ & $3.25$ $\times$ $10^{-100}$ \\\cline{2-8}
                      & Probability & $1.69$ $\times$ $10^{-112}$ & $6.25$ $\times$ $10^{-189}$ & $4.19$ $\times$ $10^{-120}$ & $6.71$ $\times$ $10^{-136}$ & $5.13$ $\times$ $10^{-71}$  & $8.72$ $\times$ $10^{-123}$ \\\cline{2-8}
                      & Stability & $2.63$ $\times$ $10^{-33}$ & $2.41$ $\times$ $10^{-194}$ & $2.78$ $\times$ $10^{-58}$ & $4.07$ $\times$ $10^{-9}$ & $1.36$ $\times$ $10^{-45}$  & $1.24$ $\times$ $10^{-97}$\\\hline
\multirow{6}{*}{RoBERTa} & Standard & $6.06$ $\times$ $10^{-100}$  & $1.82$ $\times$ $10^{-63}$ & $5.4$ $\times$ $10^{-3}$ & $3.09$ $\times$ $10^{-109}$ & $2.13$ $\times$ $10^{-57}$  & $1.15$ $\times$ $10^{-22}$\\\cline{2-8}
                      & Random & $8.38$ $\times$ $10^{-94}$ & $3.55$ $\times$ $10^{-71}$  & $3.26$ $\times$ $10^{-39}$ & $5.20$ $\times$ $10^{-101}$ & $5.12$ $\times$ $10^{-72}$  & $1.75$ $\times$ $10^{-61}$\\\cline{2-8}
                      & Probability & $5.27$ $\times$ $10^{-62}$ & $9.18$ $\times$ $10^{-193}$ & $1.39$ $\times$ $10^{-71}$ & $1.13$ $\times$ $10^{-85}$ & $4.03$ $\times$ $10^{-24}$  & $2.16$ $\times$ $10^{-72}$ \\\cline{2-8}
                      & Stability & $1.46$ $\times$ $10^{-86}$ & $3.39$ $\times$ $10^{-188}$ & $6.28$ $\times$ $10^{-5}$ & $8.71$ $\times$ $10^{-107}$ & $1.17$ $\times$ $10^{-76}$ & $1.81$ $\times$ $10^{-65}$ \\\hline
\multirow{6}{*}{XLNet} & Standard & $3.14$ $\times$ $10^{-79}$ & $4.68$ $\times$ $10^{-139}$ & $5.42$ $\times$ $10^{-112}$ & $4.17$ $\times$ $10^{-103}$ & $1.69$ $\times$ $10^{-114}$ & $5.63$ $\times$ $10^{-107}$ \\\cline{2-8}
                      & Random & $3.26$ $\times$ $10^{-71}$ & $2.97$ $\times$ $10^{-48}$ & $2.56$ $\times$ $10^{-77}$ & $5.32$ $\times$ $10^{-75}$ & $6.38$ $\times$ $10^{-32}$ & $4.38$ $\times$ $10^{-48}$ \\\cline{2-8}
                      & Probability & $4.12$ $\times$ $10^{-29}$ & $1.36$ $\times$ $10^{-63}$ & $7.25$ $\times$ $10^{-19}$ & $6.27$ $\times$ $10^{-47}$ & $1.57$ $\times$ $10^{-31}$ & $6.23$ $\times$ $10^{-32}$\\\cline{2-8}
                      & Stability & $6.17$ $\times$ $10^{-29}$ & $4.27$ $\times$ $10^{-44}$ & $1.47$ $\times$ $10^{-73}$ & $3.57$ $\times$ $10^{-41}$ & $1.79$ $\times$ $10^{-28}$ & $3.48$ $\times$ $10^{-56}$ \\\hline
\multirow{6}{*}{GPT-2} & Standard & $6.09$ $\times$ $10^{-50}$ & $1.10$ $\times$ $10^{-98}$ & $2.05$ $\times$ $10^{-57}$  & $1.22$ $\times$ $10^{-5}$ & $4.68$ $\times$ $10^{-91}$  & $1.56$ $\times$ $10^{-65}$ \\\cline{2-8}
                      & Random & $2.54$ $\times$ $10^{-22}$ & $6.97$ $\times$ $10^{-81}$ & $4.25$ $\times$ $10^{-91}$ & $9.89$ $\times$ $10^{-38}$ & $6.39$ $\times$ $10^{-77}$  & $8.70$ $\times$ $10^{-63}$ \\\cline{2-8}
                      & Probability & $5.52$ $\times$ $10^{-49}$ & $2.37$ $\times$ $10^{-89}$  & $7.02$ $\times$ $10^{-85}$ & $1.05$ $\times$ $10^{-83}$ & $1.99$ $\times$ $10^{-63}$  & $3.44$ $\times$ $10^{-49}$ \\\cline{2-8}
                      & Stability & $6.15$ $\times$ $10^{-110}$ & $3.88$ $\times$ $10^{-31}$ & $3.40$ $\times$ $10^{-66}$ & $6.27$ $\times$ $10^{-78}$ & $2.21$ $\times$ $10^{-47}$  & $2.36$ $\times$ $10^{-41}$ \\\hline
\end{tabular}
}
\label{results-statsig}
\end{table*}

\subsection{Example samples}
A few incorrectly pseudo-labeled samples from NYT-Fine dataset that are selected by probability-based selection by BERT are shown in Table~\ref{tbl:Sample sentence}
We observe a high probability assigned to each incorrect pseudo-label whereas these are learnt by the classifier at later epochs.
These wrongly annotated samples induce error that gets propagated and amplified over the iterations.
By not selecting these wrong instances, \our curbs this and boosts the performance. 

\label{sec:incorrect}
\begin{table}[t]
    \center
    \caption{Incorrectly pseudo-labeled samples selected by probability-based selection are shown below. These samples are learnt at later epochs, thus \our avoids selecting them.}
    \label{tbl:Sample sentence}
    \vspace{-3mm}
    \small
    \begin{tabular}{p{0.61\linewidth} p{0.3\linewidth}}
        \toprule
            {\textbf{Document}} & {\textbf{Pseudo-label}}\\
        \midrule
        \scriptsize Corinthians have received offer from tottenham hotspur for brazil's paulinho although the midfielder said on saturday he would not decide his future until after the confederations cup ."there is an official offer from tottenham to corinthians but, as i did when there was an inter milan offer, i'll sit and decide with my family before i make any decision," paulinho told reporters. & \thead{Football\\ Softmax Prob: 0.96 \\ Learnt Epoch: 2} \\
        \midrule
        \scriptsize Brittney griner and elena delle donne were poised to make history as the first pair of rookies from same class to start wnba all-star game. Now, neither will be playing as both are sidelined with injuries. It's a tough blow for the league, which has been marketing the two budding stars. & \thead{Baseball\\ Softmax Prob: 0.96 \\ Learnt Epoch: 2} \\
        \midrule
        \scriptsize Denmark central defender simon kjaer has joined french side lille from vfl wolfsburg on a four-year deal. Lille paid two million euros. 72 million pounds for the 24-year-old kjaer, who has won 35 caps for his country. He joined wolfsburg from palermo for 12 million euros. & \thead{Intl. Business \\ Softmax Prob: 0.94 \\Learnt Epoch: 2}\\
        \midrule
        \scriptsize Fiorentina striker giuseppe rossi is quickly making up for lost time after suffering successive knee ligament injuries which kept him out of action for the best part of two years. & \thead{Football\\ Softmax Prob: 0.95 \\Learnt Epoch: 2} \\
        \bottomrule
    \end{tabular}
\end{table}

\subsection{Learning Order vs Probability Score: Threshold Analysis}
\begin{figure}[t]
    \centering
    \includegraphics[width=1.0\linewidth]{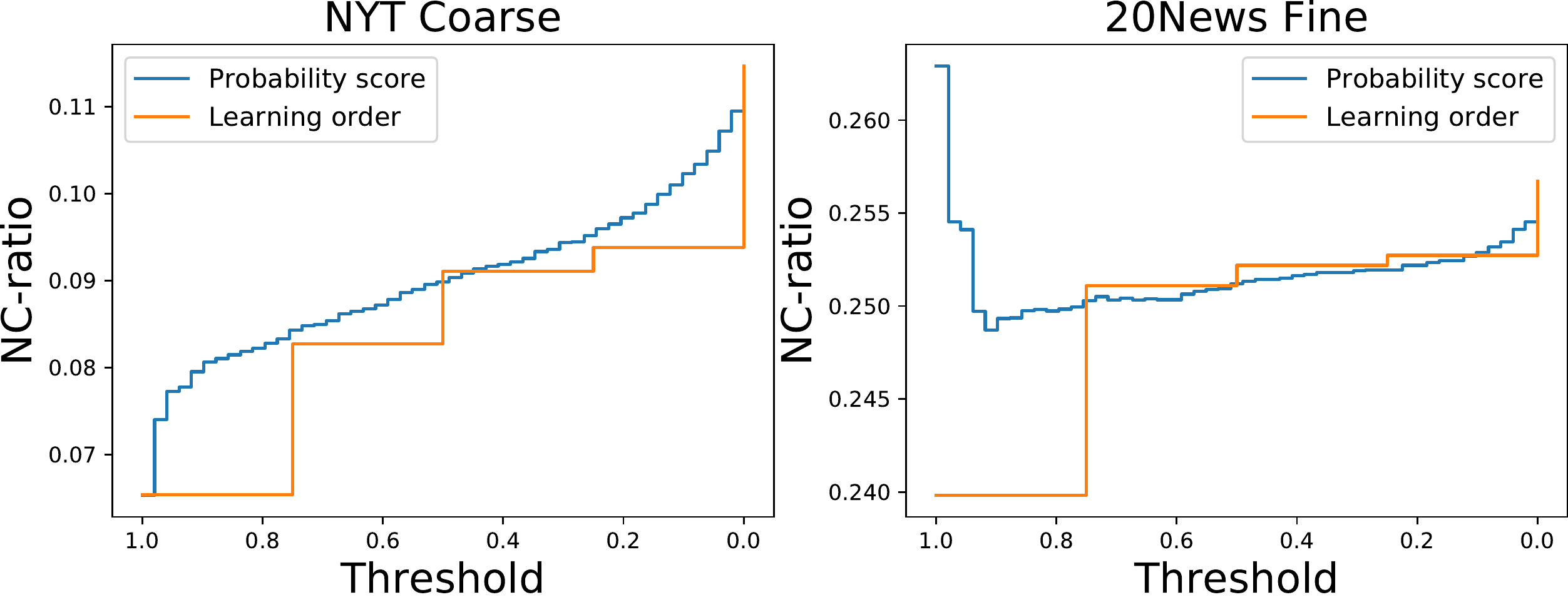}
    \vspace{-3mm}
    \caption{NC-ratios of learning order and probability score with BERT as the classifier. }
    \label{figure:nc-ratio}
    \vspace{-5mm}
\end{figure}

\begin{figure}[t]
    \subfigure[NYT Coarse]{
        \includegraphics[width=0.47\linewidth]{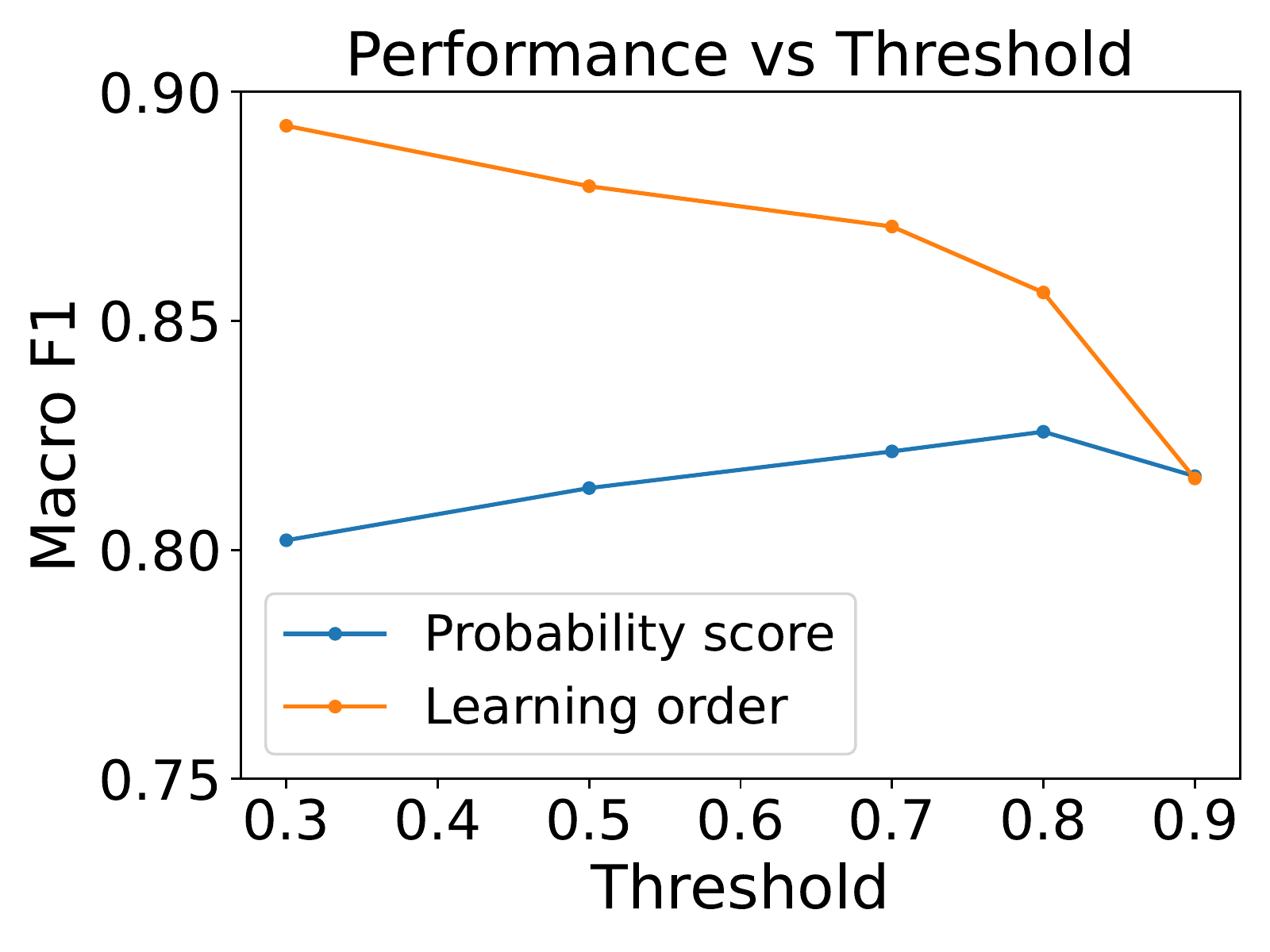}
    }
    \subfigure[20News Fine]{
        \includegraphics[width=0.47\linewidth]{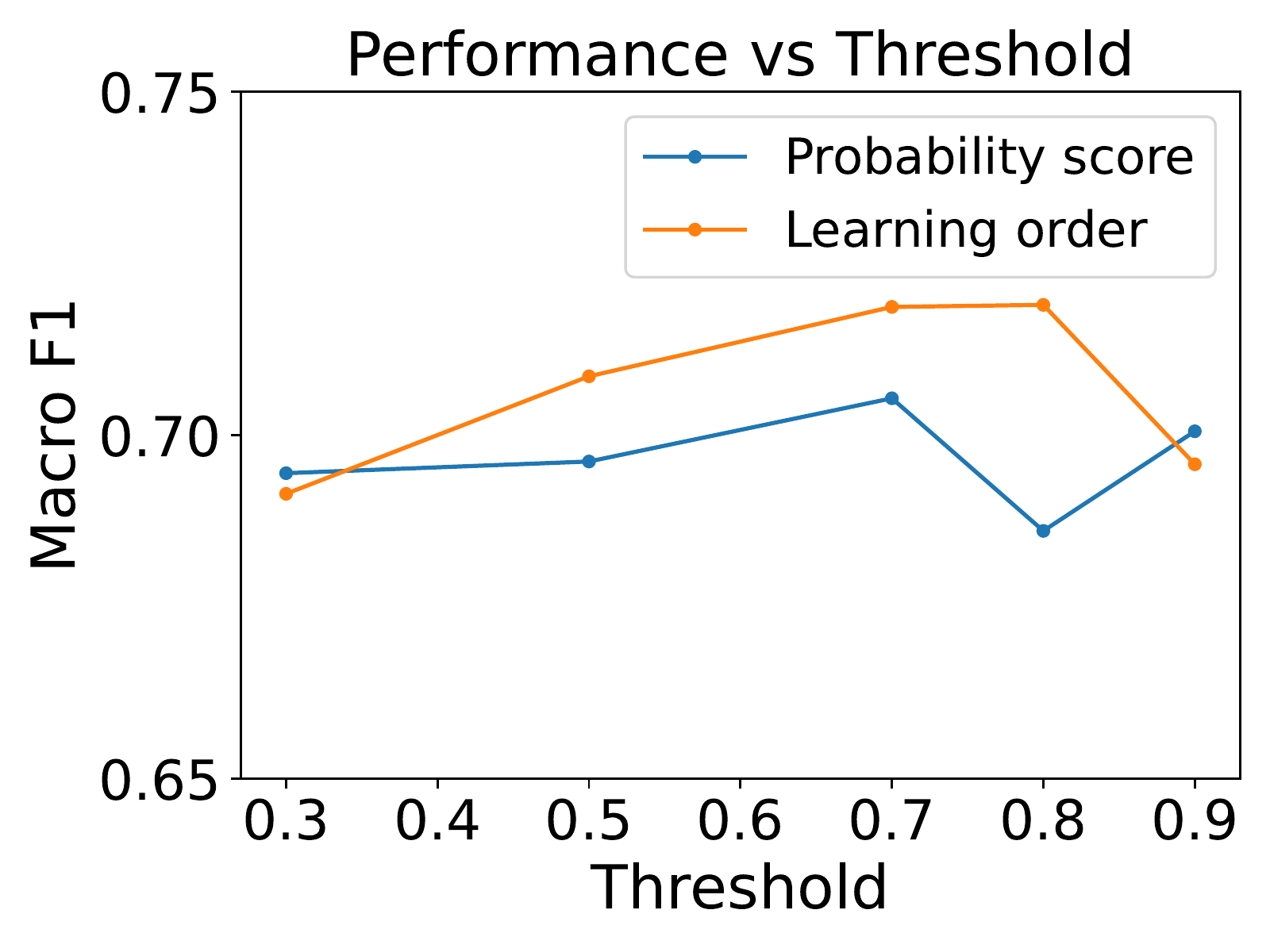}
    }
    \vspace{-3mm}
    \caption{Macro-F$_1$ scores vs Threshold on NYT-Coarse \& 20News-Fine datasets using BERT classifier with \our and Probability score based selection.}
    \label{figure:perf_thresh}
    \vspace{-5mm}
\end{figure}

Ideally, there exists a threshold for a given confidence function that perfectly distinguishes the correctly and wrongly labeled samples.
However, in practice, confidence functions may not be possible to suffice such ideal condition.
For a given confidence function, one wishes to select pseudo-labels based on a threshold such that the noise is low and the coverage is high. We define ratio between noise and coverage as \emph{NC-ratio}, namely 
$r(\kappa, \gamma) = \frac{\epsilon(\kappa, \gamma)}{\phi(\kappa, \gamma)}$.
An optimal threshold has the lowest NC-ratio.
Therefore, we evaluate confidence function by plotting NC-ratio at different thresholds.

We plot NC-ratios of learning order and probability scores with BERT classifier in Figure~\ref{figure:nc-ratio} on NYT-Coarse, 20News-Fine datasets.
To isolate them from the effects of bootstrapping, we don't perform any bootstrapping.
As shown in Figure~\ref{figure:nc-ratio}, when selecting the optimal threshold, learning order has significantly lower NC-ratios for all datasets compared to probability score. 
Furthermore, the optimal thresholds of learning order for all datasets are almost the same. In contrast, the optimal thresholds of probability score vary greatly across different datasets due to the poor calibration of DNNs. 
Finally, we also observe that the NC-ratio for probability score often changes greatly around the optimal threshold, which poses difficulty in locating the optimal threshold. 
In contrast, since there are only few possible thresholds for learning order, it is easier to find the optimal threshold.
From the performance vs threshold plot in Figure~\ref{figure:perf_thresh}, we can observe that learning order performs better than Probability score across multiple thresholds. Therefore, in terms of both performance and robustness, learning order is a more effective confidence function than probability score.

\end{document}